\documentclass{article}

\PassOptionsToPackage{numbers, compress}{natbib}

\usepackage[preprint]{neurips_2024}

\usepackage[utf8]{inputenc} 
\usepackage[T1]{fontenc}    
\usepackage{hyperref}       
\usepackage{url}            
\usepackage{booktabs}       
\usepackage{amsfonts}       
\usepackage{nicefrac}       
\usepackage{microtype}      
\usepackage{xcolor}         
\usepackage{amsmath}
\usepackage{amssymb}
\usepackage{mathtools}
\usepackage{amsthm}
\usepackage{algorithm}
\usepackage{algorithmic}
\usepackage{wrapfig}
\usepackage{array}

\title{Variational Potential Flow: A Novel Probabilistic Framework for Energy-Based Generative Modelling}

\author{
\textbf{Junn Yong Loo}$^{1}$
\quad
\textbf{Michelle Adeline}$^{1}$ 
\quad
\textbf{Arghya Pal}$^{1}$ 
\quad
\textbf{Vishnu Monn Baskaran}$^{1}$ 
\\
\textbf{Chee-Ming Ting}$^{1}$
\quad
\textbf{Rapha\"{e}l C.-W. Phan}$^{1}$
\\
$^{1}$ School of Information Technology, Monash University Malaysia\\
\texttt{\{loo.junnyong, arghya.pal, vishnu.monn\}@monash.edu}\\
\texttt{\{ting.cheeming, raphael.phan\}@monash.edu}\\
\texttt{made0008@student.monash.edu}\\
}
\theoremstyle{plain}
\newtheorem{theorem}{Theorem}
\newtheorem{proposition}[theorem]{Proposition}

\theoremstyle{definition}

\theoremstyle{remark}

\DeclareMathOperator{\mean}{\mathbb{E}}

\DeclareMathOperator{\tr}{\mathrm{tr}}
\DeclareMathOperator{\Cov}{\mathrm{Cov}}

\DeclareMathOperator{\KLD}{\mathcal{D}_\mathrm{KL}}

\begin{document}

\maketitle

\begin{abstract}
Energy based models (EBMs) are appealing for their generality and simplicity in data likelihood modeling, but have conventionally been difficult to train due to the unstable and time-consuming implicit MCMC sampling during contrastive divergence training.
In this paper, we present a novel energy-based generative framework, Variational Potential Flow (\textsf{VAPO}), that entirely dispenses with implicit MCMC sampling and does not rely on complementary latent models or cooperative training. The \textsf{VAPO} framework aims to learn a potential energy function whose gradient (flow) guides the prior samples, so that their density evolution closely follows an approximate data likelihood homotopy. An energy loss function is then formulated to minimize the Kullback-Leibler divergence between density evolution of the flow-driven prior and the data likelihood homotopy. 
Images can be generated after training the potential energy, by initializing the samples from Gaussian prior and solving the ODE governing the potential flow on a fixed time interval using generic ODE solvers. Experiment results show that the proposed \textsf{VAPO} framework is capable of generating realistic images on various image datasets. In particular, our proposed framework achieves competitive FID scores for unconditional image generation on the CIFAR-10 and CelebA datasets. 
\end{abstract}


\section{Introduction}

In recent years, deep generative modeling has garnered significant attention for unsupervised learning of complex, high-dimensional data distributions \cite{BondTaylor}. 
In particular, probabilistic generative models such as variational autoencoders \cite{VAE}, normalizing flows \cite{Rezende}, score-matching or diffusion models \cite{NCSN++,DDPM++,EDM}. Poisson flow \cite{PFGM,PFGM++}, and energy-based models (EBMs) \cite{IGEBM,JEM} aim to maximize the likelihood (probability density) underlying the data. By design, these probabilistic frameworks enhance training stability, accelerate model convergence, and reduce mode collapse compared to generative adversarial networks \cite{VEEGAN}, albeit at the cost of a slow sampling procedure and poor model scalability \cite{LSD}.
Among these frameworks, EBMs have emerged 
as a flexible and expressive class of probabilistic generative models \cite{IGEBM, LSD, IGEBM++, Shortrun_MCMC, Flow_Contrastive, Diffusion_Recovery, Cooperative_Diffusion_Recovery, JEM, SADA_JEM}. 
EBMs model 
high-dimensional data space with a network-parameterized energy potential function that assigns data regions with energy that is directly (or inversely) proportional to the unnormalized data likelihood \cite{Howto_EBM}. This provides a natural interpretation of the network model in the form of an energy landscape, thereby endowing EBMs with inherent interpretability.

Deep EBMs are particularly appealing since they impose no restrictions on the network architecture, potentially resulting in high expressiveness \cite{BondTaylor}. Moreover, they are more robust and generalize well to out-of-distribution samples \cite{IGEBM, JEM} as regions with high probability under the model but low probability under the data distribution are explicitly penalized during training. Additionally, EBMs, which trace back to Boltzmann machines \cite{Contrastive_Divergence}, have strong ties to physics models and can thus borrow insights and techniques from statistical physics for their development and analysis \cite{Feinauer}.
On these grounds, EBMs have been applied across a diverse array of applications apart from image modelling, including text generation \cite{ResidualEBM, Diffusion_Energy_Text}, point cloud synthesis \cite{GPointNet}, scene graph generation \cite{Suhail}, anomaly detection \cite{Nguyen, Yoon}, earth observation \cite{Castillo}, robot learning \cite{Planning, LEO}, trajectory prediction \cite{LB-EBM, SEEM}, and molecular design \cite{GraphEBM, Retrosynth}.

Despite a number of desirable properties, deep EBMs require implicit Langevin Markov Chain Monte Carlo (MCMC) sampling during the contrastive divergence training. MCMC sampling in a high-dimensional setting, however, has shown to be challenging due to poor mode mixing and excessively long mixing time \cite{BondTaylor, IGEBM, Shortrun_MCMC, Flow_Contrastive, JEM, MCMC_ShouldMix}. As result, energy potential functions learned with non-convergent MCMC do not have valid steady-states, in the sense that MCMC samples can differ greatly from data samples \cite{LSD}. 
Current deep EBMs are thus plagued by high variance training and high computational complexity due to MCMC sampling. 
In view of this, recent works have explored learning 
complementary latent model to amortize away the challenging MCMC sampling \cite{VAEBM, VERA, HatEBM, Latent_Prior, Diffusion_Amortized}, or cooperative learning where model-generated samples serve as initial points for subsequent MCMC revision in the latent space \cite{CoopNets, Dual_MCMC}.
While such approaches alleviate the burden of MCMC sampling, it comes at the expense of the inherent flexibility and composability of EBMs \cite{IGEBM++}.
Moreover, co-optimizing 
multiple models
adds complexity \cite{Tradeoff, Lagging} to the implementation of these approaches.


In this paper, we introduce Variational Potential Flow (\textsf{VAPO}), a novel energy-based generative framework that eliminates the need for implicit MCMC sampling and complementary models. At the core of \textsf{VAPO} lies the construction of a homotopy (smooth path) that bridges the prior distribution with the data likelihood. Subsequently, a potential flow with model-parameterized potential energy function is designed to guide the evolution of prior sample densities along this approximate data likelihood homotopy. Applying a variational approach to this path-matching strategy ultimately yields a probabilistic Poisson's equation, where the weak solution corresponds to minimizing the energy loss function of our proposed \textsf{VAPO}.

Our contributions are summarized as follows: 
\begin{itemize}

\item We introduce \textsf{VAPO}, a novel energy-based generative framework that entirely dispenses with the unstable and inefficient implicit MCMC sampling. Our proposed framework learns a potential energy function whose gradient (flow) guides the prior samples, ensuring that their density evolution path closely follows the approximate data likelihood homotopy.

\item We derive an energy loss function for \textsf{VAPO} by constructing a variational formulation of the intractable homotopy path-matching problem. Solving this energy loss objective is equivalent to minimizing the Kullback-Leibler divergence between density evolution of the flow-driven prior and the approximate data likelihood homotopy.

\item To assess the effectiveness of our proposed \textsf{VAPO} for image generation, we conduct experiments on the CIFAR-10 and CelebA datasets and benchmark the performances against state-of-the-art generative models. Our proposed framework achieves competitive FID scores of 0.0 and 0.0 for unconditional image generation on CIFAR-10 and CelebA, respectively.

\end{itemize}

\section{Background and Related Works}

In this section, we provide an overview of EBMs, particle flow, and the deep Ritz method, collectively forming the cornerstone of our proposed VAPO framework. 

\subsection{Energy-Based Models (EBMs)}

Denote $\bar{x} \in \Omega \subseteq \mathbb{R}^n$ as the training data, EBMs approximate the data likelihood ${p}_{\text{data}}(\bar{x})$ via defining a Boltzmann distribution, as follows:
\begin{align} \label{eq:Boltzmann_distribution}
\begin{split}
{p}_{\theta}(x) = \frac{e^{\Phi_{\theta}(x)}}{\int_{\Omega} e^{\Phi_{\theta}(x)} \; dx}
\end{split}
\end{align}
where $\Phi_{\theta}$ is an energy function modelled by deep neural networks.
Given that the denominator of (\ref{eq:Boltzmann_distribution}), i.e., the partition function, is analytically intractable for high-dimensional data, EBMs perform the maximum likelihood estimation (MLE) by minimizing the negative log likelihood loss $\mathcal{L}_{\mathrm{MLE}}(\theta) = \mean_{{p}_{\text{data}}(\bar{x})} [\log {p}_{\theta}(\bar{x})]$ and approximate its gradient via the contrastive divergence \cite{Contrastive_Divergence}:
\begin{align} \label{eq:contrastive_divergence}
\begin{split}
\nabla_{\theta} \mathcal{L}_{\mathrm{MLE}} = \mean_{{p}_{\text{data}}(\bar{x})}\! \big[ \nabla_{\theta} \Phi_{\theta}(\bar{x}) \big] - \mean_{{p}_{\theta}(x)}\! \big[ \nabla_{\theta} \Phi_{\theta}(x) \big]
\end{split}
\end{align}
However, EBMs are computationally intensive due to the implicit MCMC generating procedure, required for generating negative samples $x \in \Omega \sim {p}_{\theta}(x)$ for gradient computation (\ref{eq:contrastive_divergence}) during training.



\subsection{Particle Flow}

Particle flow, initially introduced by the series of papers \cite{Daum}, is a class of nonlinear Bayesian filtering (sequential inference) methods that aim to approximate the posterior distribution $p(x_{t}|\bar{x}_{0:t})$ of the state of system given the observations. While particle flow methods are closely related to normalizing flows \cite{Rezende} and neural ordinary differential equations \cite{NeuralODE}, these latter frameworks do not explicitly accommodate a Bayes update.
In particular, particle flow performs the Bayes update $p(x_{t}|\bar{x}_{0:t}) \propto p(x_{t}|\bar{x}_{0:t-1}) \, p(\bar{x}_{t}|x_{t},\bar{x}_{0:t-1})$ by subjecting prior samples $x_{t} \sim p(x_{t}|\bar{x}_{0:t-1})$ to a series of infinitesimal transformations through the ordinary differential equation (ODE) $\frac{dx}{d\tau} = v(x,\tau)$ parameterized by a flow velocity (field) function $v(x,\tau)$, in a pseudo-time interval $\tau \in [0, 1]$ in between sampling time steps. The flow velocity is designed such that the driven Kolmogorov forward path evolution (Fokker–Planck dynamics, see (\ref{eq:Kolmogorov_forward})) of the sample particles, coincides with a data log-homotopy (smooth path) that inherently perform the Bayes update. 
Despite its efficacy in time-series inference \cite{Pal, PFBR, Yang_3} and resilience to the curse of dimensionality \cite{CoD}, particle flow has yet to be explored in generative modelling for high-dimensional data.


\subsection{Deep Ritz Method}
The deep Ritz method is a deep learning-based variational numerical approach, originally proposed in \cite{Deep_Ritz}, for solving scalar elliptic partial differential equations (PDEs) in high dimensions.
Consider the following Poisson's equation, fundamental to many physical models:
\begin{align}
\begin{split} \label{eq:Dirichlet_problem}
\Delta u(x) = f(x) , \quad x \in \Omega
\end{split}
\end{align}
subject to boundary condition
\begin{align}
\begin{split} \label{eq:boundary_condition}
u(x) = 0 , \quad x \in \partial\Omega
\end{split}
\end{align}
where $\Delta$ is the Laplace operator, and $\partial \Omega$ denotes the boundary of $\Omega$. For a Sobolev function $u \in \mathcal{H}^1_0(\Omega)$ (see Proposition \ref{thm:proposition_2} for definition) and square-integrable $f \in L^2(\Omega)$, the variational principle ensures that a weak solution $u^{*}$ of the Euler-Lagrange boundary value equation (\ref{eq:Dirichlet_problem})-(\ref{eq:boundary_condition}) is equivalent to the variational problem of minimizing the Dirichlet energy \cite{Deep_Ritz_revisited}, as follows:
\begin{align}
\begin{split} \label{eq:Dirichlet_energy}
u^{*} = \arg \underset{v}{\min} \int_{\Omega} \bigg( \, \frac{1}{2} \| \nabla v(x) \|^2 - f(x) v(x) \bigg) \; dx
\end{split}
\end{align}
where $\nabla$ denotes the Del operator (gradient). In particular, the deep Ritz method parameterizes the trial energy function $v$ using neural networks, and performs the optimization (\ref{eq:Dirichlet_energy}) via stochastic gradient descent. Due to its versatility and effectiveness in handling high-dimensional PDE systems, the deep Ritz method is 
predominantly applied for finite element analysis \cite{Deep_Ritz_FEM}. In \cite{Olmez}, the deep Ritz method is used to solve the probabilistic Poisson's equation resulting from the feedback particle filter \cite{Yang_1}. 
Nonetheless, the method has not been explored for generative modelling.


\section{Variational Energy-Based Potential Flow}
In this section, we introduce a novel generative modelling framework, Variational Energy-Based Potential Flow (\textsf{VAPO}), 
drawing inspiration from both particle flow and the calculus of variations. First, we establish a homotopy that transforms a prior to the data likelihood and derive the evolution of the prior in time. Then, we design an energy-generated potential flow and a weighted Poisson's equation that aligns the evolving density distribution of transported particles with the homotopy-driven prior. Subsequently, we formulate a variational loss function where its optimization with respect to the flow-generating potential energy is equivalent to solving the Poisson's equation. Finally, we describe the model architecture that is used to parameterize the potential energy function and the backward ODE integration for generative sampling.

\subsection{Bridging Prior and Data Likelihood: Log-Homotopy Transformation} \label{ssect:LogHomotopy}
Let $\bar{x} \in \Omega$ denote the training data, ${p}_{\text{data}}(\bar{x})$ be the data likelihood, $x \in \Omega$ denote the approximate data samples.
To achieve generative modelling, our objective is to closely approximate the training data $\bar{x}$ with the data samples $x$. On this account, we define a conditional data likelihood $p(\bar{x}|x) = \mathcal{N}(\bar{x}; x, \Pi)$ with isotropic Gaussian noise with covariance $\Pi = \text{diag}(\sigma^2)$ and standard deviation $\sigma \in \Omega$. This is equivalent to considering a state space model $x = \bar{x} + \nu$, where $\nu \in \Omega \sim \mathcal{N}(\nu; 0, \Pi)$. Here, we set a small $\sigma$ so that $x$ closely resembles the training data $\bar{x}$.

Subsequently, consider a conditonal (data-conditioned) density function $\rho: \Omega^{2} \times [0, 1] \rightarrow \mathbb{R}$, as follows:
\begin{align} \label{eq:unmarginalized_homotopy_h}
\begin{split}
\rho(x; \bar{x}, t) = \frac{e^{f(x; \bar{x}, t)}}{\int_{\Omega} e^{f(x; \bar{x}, t)} \; dx}
\end{split}
\end{align}
where $f: \Omega^{2} \times [0, 1] \rightarrow \mathbb{R}$ is a log-linear function:
\begin{align} \label{eq:log-homotopy_f}
\begin{split}
f(x; \bar{x}, t) = \log q(x) + t \, \log p(\bar{x}|x)
\end{split}
\end{align}
parameterized by the auxiliary time variable $t \in [0, 1]$, and we let $q(x) = \mathcal{N}(x; 0, \Lambda)$ be a isotropic Gaussian prior density with covariance $\Lambda = \text{diag}(\omega^2)$ and standard deviation $\omega \in \Omega$. Here, $\text{diag}(\cdot)$ denotes the diagonal function.
By construction, we have 
$\rho(x; \bar{x}, 0) = q(x)$ at $t=0$, and $\rho(x; \bar{x}, 1) = p(x|\bar{x})$ at $t=1$ since we have
\begin{align} \label{eq:homotopy_t=1}
\begin{split}
\rho(x; \bar{x}, 1) &= \frac{e^{f(x; \bar{x}, 1)}}{\int_{\Omega} e^{f(x; \bar{x}, 1)} \; dx} 
= \frac{p(\bar{x}|x) \, q(x)}{{p}_{\text{data}}(\bar{x})}
= \frac{p(\bar{x}, x)}{{p}_{\text{data}}(\bar{x})}
= p(x|\bar{x})
\end{split}
\end{align}
where we have used the fact that ${p}_{\text{data}}(\bar{x}) = \int_{\Omega} e^{f(x; \bar{x}, 1)} \, dx = \int_{\Omega} p(x, \bar{x}) \, dx$.
Therefore, the conditional density function $\rho(x; \bar{x}, t)$ here (\ref{eq:unmarginalized_homotopy_h}) essentially represents a density homotopy between the prior $q(x)$ and the posterior $p(x|\bar{x})$. 

In particular, the density function $\rho(x;\bar{x},t)$ also defines a conditional (data-conditioned) homotopy between the prior $q(x)$ and the exact posterior $p(x|\bar{x})$, 
the latter of which gives a maximum a posteriori (Bayesian) estimate of the approximate data samples after observing true training data. 

To obtain an estimate of the intractable data likelihood for generative sampling, we then consider a (approximate) data likelihood homotopy 
$\bar{\rho}: \Omega \times [0, 1] \rightarrow \mathbb{R}$ as follows:
\begin{align} \label{eq:marginalized_homotopy_h}
\begin{split}
\bar{\rho}(x; t) &= \int_{\Omega} {p}_{\text{data}}(\bar{x}) \, \rho(x; \bar{x}, t) \; d\bar{x} 
\end{split}
\end{align}
Considering this, it remains that $\bar{\rho}(x; 0) = q(x)$ at $t=0$. Furthermore, given that we have
$\bar{\rho}(x; 1) = \int_{\Omega} {p}_{\text{data}}(\bar{x}) \, p(x|\bar{x}) \; d\bar{x} = \bar{p}(x)$ at $t=1$, 
the data likelihood homotopy $\bar{\rho}(x; t)$ here inherently performs a kernel density approximation of the true data likelihood, using the normalized kernel $p(x|\bar{x})$ obtained from the conditional homotopy $\rho(x;\bar{x},1)$ at $t=1$. 
Therefore, the approximate data likelihood $\bar{p}(x)$ acts as a continuous interpolation of the data likelihood ${p}_{\text{data}}(x)$, represented by Dirac delta function $\delta(x - \bar{x})$ centered on the discrete training data $\bar{x}$. 

Nevertheless, the conditional homotopy (\ref{eq:homotopy_t=1}) is intractable due to the normalizing constant in the denominator. This intractability rules out a close-form solution of the data likelihood homotopy (\ref{eq:marginalized_homotopy_h}), thus it is not possible to sample directly from the data likelihood estimate. Taking this into account, we introduce the potential flow method 
in the following section, where we model the evolution of the prior samples (particles) instead, such that their distribution adheres to the data likelihood homotopy.

\subsection{Modelling Potential Flow in a Homotopy Landscape}
Our aim is to model the flow of the prior particles in order for their distribution to follow the data likelihood homotopy and converge to the data likelihood. To accomplish this, we first derive the evolution of the latent prior density with respect to time in the following proposition.

\begin{proposition} \label{thm:proposition_1} 
Consider the data likelihood homotopy $\bar{\rho}(x; t)$ in (\ref{eq:marginalized_homotopy_h}) with Gaussian conditional data likelihood $p(\bar{x}|x) = \mathcal{N}(\bar{x}; x, \Pi)$. Then, its evolution in time $t \in [0, 1]$ is given by the following PDE:
\begin{align}
\begin{split} \label{eq:homotopy_PDE}
\frac{\partial \bar{\rho}(x;t)}{\partial t} = - \, \frac{1}{2} \, \mean_{{p}_{\text{data}}(\bar{x})} \Big[ \rho(x;\bar{x},t) \, \big( \gamma(x,\bar{x}) - \bar{\gamma}(x,\bar{x}) \big) \Big]
\end{split}
\end{align}
where 
\begin{align}
\begin{split} \label{eq:innovation}
\gamma(x,\bar{x}) = (x - \bar{x})^T \, \Pi^{-1} \, (x - \bar{x}) 
\end{split}
\end{align}
is the innovation term in the conditional data likelihood, and $\bar{\gamma}(x,\bar{x}) = \mean_{\rho(x;\bar{x},t)} [\gamma(x,\bar{x})]$ denotes the expectation of the innovation with respect to the conditional homotopy and on the latent variables.
\end{proposition}

\begin{proof}
Refer to Appendix \ref{Appendix:A}.
\end{proof}

\begin{wrapfigure}{l}{0.475\columnwidth}
\centering
\vspace{-0.25cm}
\includegraphics[width=0.475\columnwidth]{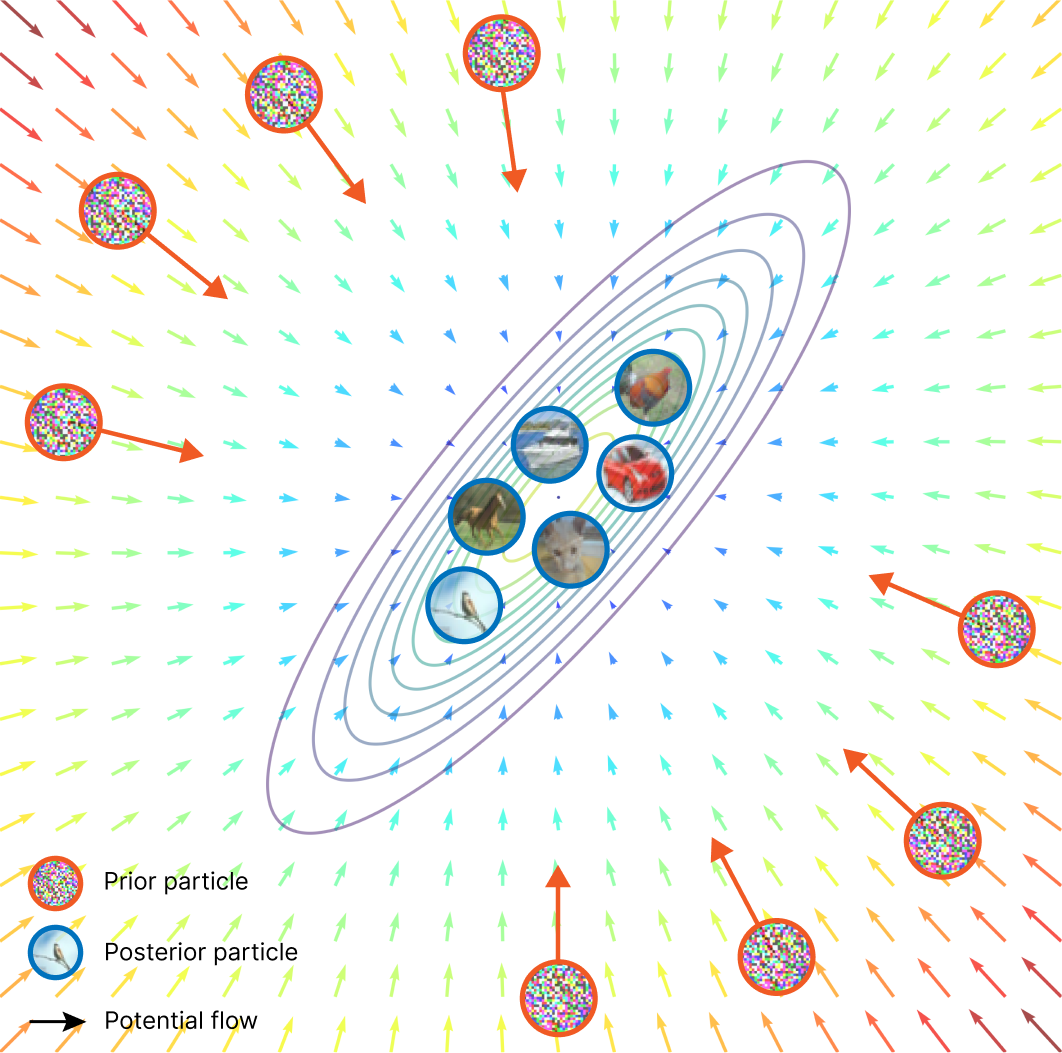}
\vspace{-0.2cm}
\caption{A planar visualization of the potential-generated field (represented by coloured arrows) that transports the prior particles towards the approximate data likelihood (represented by the blue contour).}
\label{fig:vectorfield}
\end{wrapfigure}

Our proposed potential flow method involves subjecting the latent prior samples to a potential-generated velocity field, such that the flow trajectories of these sample particles $x(t)$ within the interval $t \in [0,1]$ are governed by the following ordinary differential equation (ODE):
\begin{align} \label{eq:particle_flow_ODE}
\frac{d(x(t))}{dt} = \nabla \Phi(x(t))
\end{align}
where $\Phi: \Omega \rightarrow \mathbb{R}$ is a scalar potential energy function. Therefore, $\nabla \Phi \in \Omega$ is the velocity vector field generated by the potential energy, and $\nabla$ denotes the Del operator (gradient) with respect to the data samples $x$. The scalar potential (\ref{eq:particle_flow_ODE}) is a result of the Helmholtz decomposition of a vector field $V(x(t))$ and disregarding the solenoidal (rotational) component.
Henceforth, the time variable of $x(t)$ is implicitly assumed, thus we omit it for simplicity.

Considering a potential flow of the form (\ref{eq:particle_flow_ODE}), a direct consequence is that the (approximate likelihood) density ${\rho}^{\Phi}$ of the flow-driven prior samples evolves according to a Fokker–Planck (Kolmogorov forward) equation as follows:
\begin{align}
\begin{split} \label{eq:Kolmogorov_forward}
\frac{\partial {\rho}^{\Phi}(x;t)}{\partial t} = - \, \nabla \cdot \Big( {\rho}^{\Phi}(x;t) \, \nabla \Phi(x) \Big)
\end{split}
\end{align}
where $\nabla \cdot$ denotes the divergence operator. 
In particular, the Fokker–Planck equation (\ref{eq:Kolmogorov_forward}) exemplifies a form of continuity (transport) equation commonly used for modelling fluid advection. In this analogy, ${\rho}^{\Phi}$ corresponds to fluid density, $\Phi$ represents a field-driving potential energy, and its gradient $\nabla \Phi$ acts as the resulting conservative (irrotational) velocity field.

The goal of our proposed framework is to model the potential energy function in the potential flow (\ref{eq:particle_flow_ODE}), such that the progression of the prior density subject to potential flow emulates the evolution of the data likelihood homotopy.
In particular, we seek to solve the problem of minimizing the Kullback-Leibler divergence (KLD) in the following proposition.

\begin{proposition} \label{thm:proposition_2}
Consider a potential flow of the form (\ref{eq:particle_flow_ODE}) and given that $\Phi \in \mathcal{H}^1_0(\Omega, \bar{\rho})$, where $\mathcal{H}^n_0$ denotes the (Sobolev) space of $n$-times differentiable functions that are compactly supported, and square-integrable with respect to data likelihood homotopy $\bar{\rho}(x;t)$. 

Then, the problem of solving for the optimal potential energy function $\Phi(x)$ that satisfies the following probabilistic (density-weighted) Poisson's equation:
\begin{align}
\begin{split} \label{eq:unmarginalized_PDE_equation}
& \nabla \cdot \Big( \bar{\rho}(x;t) \, \nabla \Phi(x) \Big) 
= \frac{1}{2} \, \mean_{{p}_{\text{data}}(\bar{x})} \Big[ \rho(x;\bar{x},t) \, \big( \gamma(x,\bar{x}) - \bar{\gamma}(x,\bar{x}) \big) \Big]
\end{split}
\end{align}
is equivalent to minimizing the KLD $\KLD \big[ {\rho}^{\Phi}(x;t) \,\|\, \bar{\rho}(x;t) \big]$ between the flow-driven prior ${\rho}^{\Phi}(x;t)$ and the data likelihood homotopy $\bar{\rho}(x;t)$ at time $t$.
\end{proposition}

\begin{proof}
Refer to Appendix \ref{Appendix:B}.
\end{proof}

In hindsight, the left-hand side of the probabilistic Poisson's equation (\ref{eq:unmarginalized_PDE_equation}) resembles the evolution of the flow-driven prior given by the Fokker-Plank equation (\ref{eq:Kolmogorov_forward}). In addition, the right-hand side resembles the evolution of data likelihood homotopy given by PDE (\ref{eq:homotopy_PDE}), with the conditional homotopy $\rho(x;\bar{x},t)$ replaced by flow-driven prior ${\rho}^{\Phi}(x;t)$. Therefore, the probabilistic Poisson's equation is an attempt to solve the approximation
$\frac{\partial {\rho}^{\Phi}(x;t)}{\partial t} \equiv \frac{\partial \bar{\rho}(x;t)}{\partial t}$.

Nevertheless, explicitly solving the probabilistic Poisson's equation (\ref{eq:unmarginalized_PDE_equation}) is challenging in a high-dimensional setting. Numerical methods that approximate the solution often do not scale well with the data dimension. For example, the Galerkin approximation requires a selection of the basis functions, which becomes non-trivial when the dimensionality is high \cite{Yang_2}. The diffusion map-based algorithm, on the other hand,
requires a large number of particles, which grows exponentially with respect to the dimensionality, in order to achieve error convergence \cite{Diffusion_Map}. Taking this into consideration, we propose an energy loss function in the following section, where we cast the Poisson's equation as a variational problem compatible with stochastic gradient descent.

\subsection{Variational Energy Loss Function Formulation: Deep Ritz Approach}

In this section, we introduce an energy method which presents a variational formulation of the probabilistic Poisson's equation.
Given that the aim is to minimize the divergence between the data likelihood homotopy and the flow-driven prior and directly solving the probabilistic Poisson's equation is difficult, we first consider a weak formulation of (\ref{eq:unmarginalized_PDE_equation}) as follows:
\begin{align}
\begin{split} \label{eq:weak_formulation}
\int_{\Omega} \bigg( &\, \frac{1}{2} \, \mean_{{p}_{\text{data}}(\bar{x})} \Big[ \rho(x;\bar{x},t) \, \big( \gamma(x,\bar{x}) - \bar{\gamma}(x,\bar{x}) \big) \Big] 
- \nabla \cdot \Big( \bar{\rho}(x;t) \, \nabla \Phi(x) \Big) \bigg) \, \Psi(x) \; dx \,=\, 0
\end{split}
\end{align}
where the equation must hold for all differentiable trial functions $\Psi$. In the following proposition, we introduce an energy loss objective that is equivalent to solving this weak formulation of the probabilistic Poisson's equation.

\begin{proposition} \label{thm:proposition_3}
The variational problem of minimizing the following loss function:
\begin{align} \label{eq:variational_functional}
\begin{split}
\mathcal{L}(\Phi; t) = &\; \frac{1}{2} \, \Cov_{\rho(x;\bar{x},t) \, {p}_{\text{data}}(\bar{x})} \big[ \Phi(x) , \gamma(x,\bar{x}) \big] 
+ \frac{1}{2} \, \mean_{\bar{\rho}(x;t)} \Big[ \big\| \nabla \Phi(x) \big\|^{2} \Big]
\end{split}
\end{align}
with respect to the potential energy $\Phi$, 
is equivalent to solving the weak formulation (\ref{eq:weak_formulation}) of the probabilistic Poisson's equation (\ref{eq:unmarginalized_PDE_equation}).
Here, $\|\cdot\|$ denotes the Euclidean norm, and $\Cov$ denotes the covariance.

Furthermore, the variational problem (\ref{eq:variational_functional}) has a unique solution if for all energy functions $\Phi \in \mathcal{H}^1_0(\Omega; \bar{\rho})$, the data likelihood homotopy $\bar{\rho}$ satisfy the Poincar\'e inequality:
\begin{align} \label{eq:Poincare_inequality}
\begin{split}
\mean_{\bar{\rho}(x;t)} \Big[ \big\| \nabla \Phi(x) \big\|^{2} \Big] \geq \lambda \, \mean_{\bar{\rho}(x;t)} \Big[ \big\| \Phi(x) \big\|^{2} \Big]
\end{split}
\end{align}
for some positive scalar constant $\lambda > 0$ (spectral gap).
\end{proposition}

\begin{proof}
Refer to Appendix \ref{Appendix:C}.
\end{proof}

In sum, leveraging Propositions \ref{thm:proposition_2} and \ref{thm:proposition_3}, we reformulate the intractable task of minimizing the KLD between flow-driven prior and data likelihood homotopy equivalently as a variational problem with energy loss function (\ref{eq:variational_functional}). By optimizing the potential energy function $\Phi$ with respect to the energy loss and transporting the prior samples through the potential flow ODE (\ref{eq:particle_flow_ODE}), the prior particles follow a trajectory that accurately approximates the data likelihood homotopy. In doing so, the potential flow $\nabla \Phi$ drives the prior samples to posterior regions densely populated with data, thus enabling us to perform generative modelling.

The minimum covariance objective in (\ref{eq:variational_functional}) plays an important role 
by ensuring that the normalized innovation is inversely proportional to the potential energy. As a result, the potential-generated velocity field $\nabla \Phi$ consistently points in the direction of greatest potential ascent, thereby driving the flow of prior particles towards high likelihood regions of the true posterior, as illustrated in Figure \ref{fig:vectorfield}. In other words, the potential energy is conjugate to the approximate data likelihood $\bar{p}(\bar{x})$, analogous to Hamiltonian fluid mechanics \cite{Hamiltonian}. It is worth noticing that instead of being an ad hoc addition, the L2 regularization term $\mean_{\bar{\rho}(x;t)} [ \| \nabla \Phi(x) \|^{2} ]$ on the velocity field in (\ref{eq:variational_functional}) arises, from first-principle derivation, as a direct consequence of considering the data likelihood homotopy.

Given that the aim is to solve the probabilistic Poisson's equation (\ref{eq:unmarginalized_PDE_equation}) for all $t$, we include an auxiliary time integral to the energy loss function (\ref{eq:variational_functional}) as follows:
\begin{align} \label{eq:auxiliary_time_integral}
\begin{split}
\mathcal{L}^{\mathrm{VAPO}}(\theta) = \int_{\mathbb{R}} \, \mathcal{L}(\theta; t) \; dt = \mean_{\,\mathcal{U}(t;0,1)} \big[ \mathcal{L}(\theta; t) \big]
\end{split}
\end{align}
where we have applied Monte Carlo integration, and $\mathcal{U}(a,b)$ denotes the uniform distribution over interval $[a,b]$.
In addition, the data likelihood homotopy may not satisfy the Poincaré inequality (\ref{eq:Poincare_inequality}). Hence, we include the right-hand side of the inequality to the loss function (\ref{eq:variational_functional}) to enforce uniqueness of its minimizer. This addtional L2 loss also regularize the energy function, preventing its values from exploding. The spectral gap constant $\lambda$ is left as a training hyperparameter.

In addition, the energy loss (\ref{eq:variational_functional}) requires us to sample from the conditional and data likelihood density homotopies. By design, both the prior $q(x) = \mathcal{N}(x; 0, \Lambda)$ and the conditional data likelihood $p(\bar{x}|x) = \mathcal{N}(\bar{x}; x, \Pi)$ are assumed to be Gaussian. As a consequence, the Bayes update (\ref{eq:unmarginalized_homotopy_h}) results in a Gaussian density $\rho(x;\bar{x},t) = \mathcal{N}\big(x; \mu(\bar{x},t), \Sigma(\bar{x},t)\big)$, 
from which the time-varying mean and covariance can be derived using the Bayes' theorem \cite{PRML}, as follows:
\begin{align} \label{eq:unmarginalized_homotopy_statistics}
&\mu(\bar{x},t) = t \, \Sigma(t) \, \Pi^{-1} \, \bar{x}, \qquad
\Sigma(t) = \big(\Lambda^{-1} \,+\, t \, \Pi^{-1}\big)^{-1}
\end{align}
Therefore, to sample from $\rho(x;\bar{x},t)$ or $\rho(x;t)$, we first sample data $x$ from ${p}_{\text{data}}(\bar{x})$ and compute the mean and covariance according to (\ref{eq:unmarginalized_homotopy_statistics}). Then, we can generate samples of the approximate data $x$ using the reparameterization trick $x = \mu(\bar{x},t) + \sqrt{\Sigma}(t) \, \epsilon$, where $\epsilon \sim \mathcal{N}(\epsilon; 0,I)$ and $\sqrt{\Sigma}$ is the square root decomposition of $\Sigma$, i.e., $\Sigma = \sqrt{\Sigma}\sqrt{\Sigma}^T$. A detailed derivation of (\ref{eq:unmarginalized_homotopy_statistics}) is provided in Appendix \ref{Appendix:D}.

Nevertheless, parameterizing the conditional homotopy using mean and covariance (\ref{eq:unmarginalized_homotopy_statistics}) causes it to converge too quickly to the posterior $\rho(x;\bar{x},1) = p(x|\bar{x})$. As a consequence, most samples are closely clustered around the observed data. To mitigate this issue, a strategy is to slow down its convergence by reparameterizing it with $t + \varepsilon = e^{\tau}$, where $\tau \in [\ln \varepsilon, \ln (1+\varepsilon)]$. This time reparameterization compels $t+\varepsilon$ to follow a log-uniform (reciprocal) distribution $\mathcal{R}(t+\varepsilon;\varepsilon,1+\varepsilon)$ defined over the interval $[\varepsilon,1+\varepsilon]$. Here, the hyperparameter $\varepsilon$ is a small positive constant that determines the sharpness of the log-uniform density, and the rate at which its tail decays to zero. 

Incorporating all of the above considerations, the final energy loss function becomes:
\begin{align} \label{eq:VAPO_loss}
\begin{split}
\mathcal{L}_{\mathrm{VAPO}}(\theta) &= \frac{1}{2} \, \mean_{\mathcal{R}(t+\varepsilon;\,\varepsilon,1+\varepsilon)} \big[ \mathcal{L}(\theta;t) \big]
\end{split}
\end{align}
where
\begin{align} \label{eq:VAPO_loss_1}
\begin{split}
\mathcal{L}(\theta; t) = &\; \Cov_{\rho(x;\bar{x},t) \, {p}_{\text{data}}(\bar{x})} \big[ \Phi_{\theta}(x) , \gamma(x,\bar{x}) \big] \\
&+ \mean_{\bar{\rho}(x;t)} \Big[ \| \nabla \Phi_\theta(x) \big\|^{2} \Big] 
+ \lambda \mean_{\bar{\rho}(x;t)} \Big[ \big\| \Phi_{\theta}(x) \big\|^{2} \Big]
\end{split}
\end{align}
The algorithm for training \textsf{VAPO} is shown in Algorithm \ref{algo:VAPO_training}.

\begin{algorithm}[t]
\caption{\textsf{VAPO} Training}
\label{algo:VAPO_training}
\begin{algorithmic}
\STATE {\bfseries Input:} Initial energy model $\Phi_\theta$, spectral gap constant $\lambda$, sharpness constant $\varepsilon$, standard deviation $\omega$ of prior density, standard deviation $\sigma$ of conditional data likelihood, and batch size $B$.
\REPEAT
\STATE Sample observed data $\bar{x}_i \sim {p}_{\text{data}}(\bar{x})$, $t_i \sim \mathcal{R}(t+\varepsilon;\,\varepsilon,1+\varepsilon)$, and $\epsilon_i \sim \mathcal{N}(\epsilon; 0,I)$
\STATE Sample $x_i \sim \rho(x;\bar{x},t)$ via reparameterization $x_i = \mu(\bar{x}_i,t_i) + \sqrt{\Sigma}(t_i) \, \epsilon_i$ based on (\ref{eq:unmarginalized_homotopy_statistics})
\STATE Compute gradient $\nabla \Phi_{\theta}(x_i)$ w.r.t. $x_i$ via backpropagation
\STATE Calculate innovation $\gamma(x_i,\bar{x}_i)$ based on (\ref{eq:innovation})
\STATE Calculate \textsf{VAPO} loss $\mathcal{L}_{\mathrm{VAPO}}(\theta) = \frac{1}{B} \sum_{i=1}^{B} \mathcal{L}(\theta;t_i)$ based on (\ref{eq:VAPO_loss})-(\ref{eq:VAPO_loss_1})
\STATE Update energy model parameters $\theta$ with the gradient of $\mathcal{L}_{\mathrm{VAPO}}(\theta)$
\UNTIL{$\theta$ converged}
\end{algorithmic}
\end{algorithm}

\subsection{Energy Parameterization and ODE Sampling}

To implement stochastic gradient descent on top of the energy loss function (\ref{eq:auxiliary_time_integral}), we adopt the deep Ritz approach and in particular, we model 
the potential energy function $\Phi_{\theta}$ as deep neural networks with parameters $\theta$. Here, we restrict our model architecture to convolutional and fully-connected layers, which are shown to satisfy the universal approximation property within weighted Sobolev spaces, i.e., the neural network model densely approximates functions in $\mathcal{H}^1_0(\Omega, \bar{\rho})$ and enables model convergence. 



\newcolumntype{P}[1]{>{\centering\arraybackslash}p{#1}}
\begin{table}[t]
\centering
\caption{Comparison of FID scores on unconditional CIFAR-10 image generation. FID baselines are obtained from \cite{Cooperative_Diffusion_Recovery}.}
\label{tab:fid_cifar10}
\centering
\setlength{\aboverulesep}{0pt}
\setlength{\belowrulesep}{1pt}
\begin{tabular}{p{0.3\textwidth}p{0.06\textwidth}|p{0.3\textwidth}p{0.06\textwidth}}
\toprule
\textbf{Models} & \textbf{FID $\downarrow$} & \textbf{Models} & \textbf{FID $\downarrow$} \\
\midrule
\multicolumn{2}{c|}{\textbf{\textit{EBM-based methods}}} & \multicolumn{2}{c}{\textbf{\textit{Other likelihood-based methods}}} \\
\midrule
EBM-SR \cite{Shortrun_MCMC} &  44.5 & VAE \cite{VAE}  & 78.4 \\
JEM \cite{JEM} & 38.4 & PixelCNN \cite{pixelcnn}   & 65.9 \\
EBM-IG \cite{IGEBM} & 38.2 & PixelIQN \cite{pixeliqn} & 49.5  \\
EBM-FCE \cite{Flow_Contrastive} & 37.3 & ResidualFlow \cite{residualflow}  &  47.4 \\
CoopVAEBM \cite{CoopVAEBM} & 36.2 &  Glow \cite{glow} & 46.0  \\
CoopNets \cite{CoopNets} & 33.6 & DC-VAE \cite{dc-vae} & 17.9  \\
\cmidrule(r){3-4}
Divergence Triangle \cite{divergence_triangle} & 30.1 & \multicolumn{2}{c}{\textbf{\textit{GAN-based methods}}}  \\
\cmidrule(r){3-4}
VERA \cite{VERA} & 27.5 & WGAN-GP \cite{WGAN}   & 36.4 \\
EBM-CD \cite{IGEBM++}  & 25.1 & SN-GAN \cite{Miyato}   & 21.7  \\
GEBM \cite{GEBM}  & 19.3 & SNGAN-DDLS \cite{HatEBM} & 15.4  \\
HAT-EBM \cite{HatEBM} & 19.3 & 
BigGAN \cite{BigGAN} & 14.8 \\
\cmidrule(r){3-4}
CF-EBM \cite{CFEBM} & 16.7 & \multicolumn{2}{c}{\textbf{\textit{Score-based and Diffusion methods}}} \\
\cmidrule(r){3-4}
CoopFlow \cite{CoopFlow} & 15.8 & NCSN \cite{NCSN} & 25.3 \\
CLEL-base \cite{CLEL} & 15.3 & NCSN-v2 \cite{NCSNv2} & 10.9 \\
VAEBM \cite{VAEBM} & 12.2 &  DDPM Distil. \cite{DDPM_distil} & 9.36 \\
DRL \cite{Diffusion_Recovery} & 9.58 & DDPM \cite{DDPM} & 3.17 \\
\cmidrule(l){1-2}
\textsf{VAPO} (Ours) & \textbf{16.6} & NCSN++\cite{NCSN++} & 2.20 \\
\bottomrule
\end{tabular}
\vspace{-0.4cm}
\end{table}

\begin{wraptable}{r}{0.35\columnwidth}
\vspace{-0.925cm}
\caption{Comparison of FID scores on unconditional CelebA $64 \times 64$. FID baselines obtained from \cite{Diffusion_Recovery}.}
\label{tab:fid_celeba}
\vspace{+0.2cm}
\centering
\setlength{\aboverulesep}{0pt}
\setlength{\belowrulesep}{1pt}
\begin{tabular}{p{0.2\textwidth}p{0.06\textwidth}}
\toprule
\textbf{Models} & \textbf{FID $\downarrow$} \\ \midrule
NCSN \cite{NCSN} & 25.3 \\
NCSN-v2 \cite{NCSNv2} & 10.2 \\ \midrule
EBM-Triangle \cite{EBM-Triangle} & 24.7 \\
EBM-SR \cite{Shortrun_MCMC} & 23.0 \\
Divergence Triangle \cite{divergence_triangle} & 18.2 \\
CoopNets \cite{CoopNets} & 16.7 \\
\midrule
\textsf{VAPO} (Ours) & \textbf{14.5} \\ \bottomrule
\end{tabular}
\vspace{-1cm}
\end{wraptable}

After modelling the potential energy $\Phi_{\theta}$, the gradient $\nabla \Phi_\theta$ can be used to generate approximate data samples from the potential flow ODE (\ref{eq:particle_flow_ODE}). Given that the ODE is defined on the interval $[0,1]$ by construction, its starting and terminal time is predetermined and therefore known, in contrast to %
most flow-based generative frameworks. On top of that, the potential flow ODE is compatible with general-purpose ODE solvers, such as the explicit and implicit Runge-Kutta methods of different orders and the forward and backward Euler methods, which can readily be employed for sampling.

\section{Experiments}

In this section, we show that \textsf{VAPO} is an effective generative model for images. In particular, Section 4.1 demonstrates that \textsf{VAPO} is capable of generating realistic unconditional images on the well-known CIFAR-10 and CelebA datasets. Section 4.2 demonstrates that \textsf{VAPO} is capable of performing smooth interpolation between two generated samples.
Implementation details, including model architecture and training, numerical ODE solver, datasets and FID evaluation are provided in Appendix \ref{sec:experimental_details}.
Apart from that, we also show that \textsf{VAPO} exhibits extensive mode coverage and robustness to anomalous data, as well as generalizing well to unseen data without over-fitting. Specifically, Appendix \ref{sec:mode_evaluation}.1 evaluates model over-fitting and generalization based on the energy histogram of CIFAR-10 train and test sets and the nearest neighbors of generated samples. Appendix \ref{sec:mode_evaluation}.2 examines robustness to anomalous data by assessing its performance on out-of-distribution (OOD) detection 
on various image datasets.


\begin{figure}[t]
\centering
\includegraphics[width=\columnwidth]{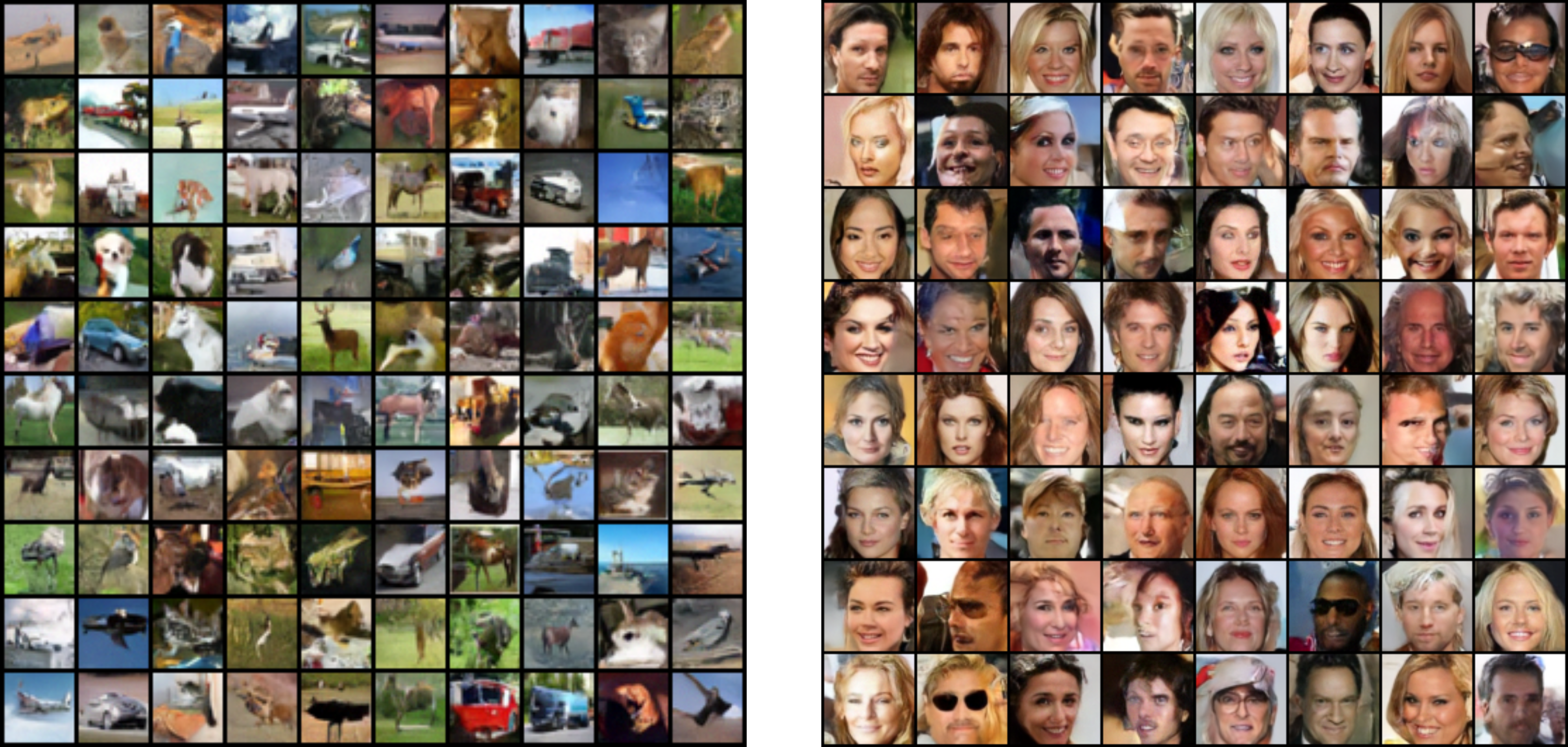}
\vspace{-0.4cm}
\caption{Generated samples on unconditional CIFAR-10 $32\times 32$ (left) and CelebA $64\times 64$ (right).}
\label{fig:cifar10_small}
\vspace{-0.4cm}
\end{figure}

\subsection{Unconditional Image Generation}

Figure \ref{fig:cifar10_small} shows the uncurated and unconditional image samples generated from the learned energy model on the CIFAR-10 and CelebA $64 \times 64$ datasets. More generated samples are provided in Appendix \ref{sec:additional_results}. The samples are of decent quality and resemble the original datasets despite not having the highest fidelity as achieved by state-of-the-art models. Tables 1 and 2 summarize the quantitative evaluations of our proposed \textsf{VAPO} model in terms of FID \cite{FID} scores on the CIFAR-10 and CelebA datasets. On CIFAR-10, \textsf{VAPO} achieves a competitive FID that is better than the majority of existing EBM-based generative models. Having dispensed with the implicit MCMC sampling, \textsf{VAPO} still outperforms most of the EBM approaches without relying on complementary latent models or cooperative training. 
On CelebA, \textsf{VAPO} obtains an FID that outperforms some existing EBMs but falls short compared to \cite{NCSNv2} and state-of-the-art models.

\subsection{Image Interpolation}

\begin{figure}[ht]
\centering
\includegraphics[width=\columnwidth]{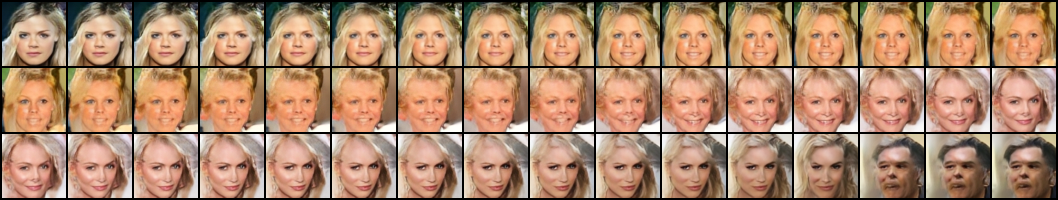}
\vspace{-0.4cm}
\caption{Interpolation results between the leftmost and rightmost generated CelebA $64 \times 64$ samples.}
\label{fig:celeba_interp_small}
\vspace{-0.4cm}
\end{figure}

Figure \ref{fig:celeba_interp_small} shows the interpolation results between pairs of generated CelebA samples, where it demonstrates that \textsf{VAPO} is capable of smooth and semantically meaningful image interpolation. To perform interpolation for two samples $x_1(1)$ and $x_2(1)$, we construct a spherical interpolation between the initial Gaussian noise $x_1(0)$ and $x_2(0)$, and subject them to sampling over the potential flow ODE. More interpolation results on CIFAR-10 and CelebA are provided in Appendix \ref{sec:additional_results}.

\section{Limitations and Future Work}

We propose \textsf{VAPO}, a novel energy-based generative modelling framework without the need for expensive and unstable MCMC runs amidst training. Despite the improvement over the majority of existing EBMs, there is still a large performance gap between \textsf{VAPO} and the state-of-the-art score-based (or diffusion) and Poisson flow models \cite{NCSN++, DDPM++, PFGM}. To close this gap, diffusion recovery likelihood \cite{Diffusion_Recovery, Cooperative_Diffusion_Recovery}, which is shown to be more tractable than marginal likelihood, can be incorporated into the \textsf{VAPO} framework for a more controlled diffusion-guided energy optimization. The dimensionality augmentation technique of \cite{PFGM, PFGM++} can also be integrated given that fundamentally, both Poisson flow and \textsf{VAPO} aim to model potential field governed by a Poisson's equation. On top of that, the scalability of \textsf{VAPO} to higher resolution images and its generalizability to other data modalities have yet to be validated. In addition, the current \textsf{VAPO} framework does not allow for class-conditional generation. Moreover, the training of \textsf{VAPO} requires a large number of iterations to converge and thus warrants improvement. These important aspects are earmarked for future extensions of our work.


\newpage
\bibliographystyle{unsrt}
\bibliography{refs}

\newpage
\appendix

\section{Proofs and Derivations}
\label{sec:proofs_and_derivations}

\subsection{Proof of Proposition \ref{thm:proposition_1}} \label{Appendix:A}
\begin{proof}
Differentiating the conditional homotopy $\rho(x;\bar{x},t)$ in (\ref{eq:unmarginalized_homotopy_h}) with respect to $t$, we have
\begin{align} \label{eq:homotopy_derivative}
\begin{split}
&\frac{\partial \rho(x;\bar{x},t)}{\partial t} 
= \frac{1}{\int_{\Omega} e^{f(x;\bar{x},t)} \; dx} \, \frac{\partial [e^{f(x;\bar{x},t)}]}{\partial t}
- \frac{e^{f(x;\bar{x},t)}}{[\int_{\Omega} e^{f(x;\bar{x},t)} \; dx]^2} \, \frac{\partial [\int_{\Omega} e^{f(x;\bar{x},t)} \; dx]}{\partial t} \\
&= \frac{1}{\int_{\Omega} e^{f(x;\bar{x},t)} \; dx} \, \frac{\partial [e^{f(x;\bar{x},t)}]}{\partial f} \, \frac{\partial f(x;\bar{x},t)}{\partial t}
- \frac{e^{f(x;\bar{x},t)}}{[\int_{\Omega} e^{f(x;\bar{x},t)} \; dx]^2} \, \int_{\Omega} \frac{\partial [e^{f(x;\bar{x},t)}]}{\partial f} \, \frac{\partial f(x;\bar{x},t)}{\partial t} \; dx \\
&= \frac{e^{f(x;\bar{x},t)}}{\int_{\Omega} e^{f(x;\bar{x},t)} \; dx} \, \frac{\partial f(x;\bar{x},t)}{\partial t}
- \frac{e^{f(x;\bar{x},t)}}{\int_{\Omega} e^{f(x;\bar{x},t)} \; dx} \, \int_{\Omega} \frac{e^{f(x;\bar{x},t)}}{\int_{\Omega} e^{f(x;\bar{x},t)} \; dx} \, \frac{\partial f(x;\bar{x},t)}{\partial t} \; dx \\
&= \rho(x;\bar{x},t) \left( \frac{\partial f(x;\bar{x},t)}{\partial t} - \int_{\Omega} \rho(x;\bar{x},t) \, \frac{\partial f(x;\bar{x},t)}{\partial t} \; dx \right) \\
&= - \, \frac{1}{2} \, \rho(x;\bar{x},t) \Bigg( (x - \bar{x})^T \, \Pi^{-1} \, (x - \bar{x}) - \int_{\Omega} \rho(x;\bar{x},t) \, (x - \bar{x})^T \, \Pi^{-1} \, (x - \bar{x}) \; dx \Bigg)
\end{split}
\end{align}
where we used the quotient rule in the first equation, and chain rule in the second equation.

Let $\gamma(x,\bar{x}) = (x - \bar{x})^T \, \Pi^{-1} \, (x - \bar{x})$ and using the fact that
\begin{align} \label{eq:marginalized_homotopy_derivative}
\begin{split}
\frac{\partial \bar{\rho}(x;t)}{\partial t} = \frac{\partial \int_{\Omega} \rho(x;\bar{x},t) \, {p}_{\text{data}}(\bar{x}) \; d\bar{x}}{\partial t} = \int_{\Omega} \frac{\partial \rho(x;\bar{x},t)}{\partial t} \, {p}_{\text{data}}(\bar{x}) \; d\bar{x}
\end{split}
\end{align}
we can substitute (\ref{eq:homotopy_derivative}) into (\ref{eq:marginalized_homotopy_derivative}) to get
\begin{align} \label{eq:marginalized_homotopy_derivative_expand}
\begin{split}
\frac{\partial \bar{\rho}(x;t)}{\partial t} = - \, \frac{1}{2} \int_{\Omega} {p}_{\text{data}}(\bar{x}) \, \rho(x;\bar{x},t) \, \bigg( \gamma(x,\bar{x}) - \int_{\Omega} \rho(x;\bar{x},t) \, \gamma(x,\bar{x}) \; dx \bigg) \; d\bar{x}
\end{split}
\end{align}
Given that both $\rho(x;\bar{x},t)$ and ${p}_{\text{data}}(x)$ are normalized (proper) density functions, writing (\ref{eq:marginalized_homotopy_derivative_expand}) in terms of expectations
yields the PDE in (\ref{eq:homotopy_PDE}). 
\end{proof}


\subsection{Proof of Proposition \ref{thm:proposition_2}} \label{Appendix:B}
Here, we used the Einstein tensor notation interchangeably with the conventional notation for vector dot product and matrix-vector multiplication in PDE. 

Given that the context is clear, we write $g_t(\cdot)$ in place of time-varying functions $g(\cdot, t)$. For brevity, we will also omit the time index $t$, and write $g$ in place of $g_t(x)$. 

\begin{proof}
Applying the forward Euler method to the particle flow ODE (\ref{eq:particle_flow_ODE}) using step size $\Delta_t$, we obtain:
\begin{align} 
\begin{split} \label{eq:forward_Euler_rv}
x_{t+\Delta_t} &= \alpha_t(x_t) = x_t + \Delta_t \, u(x_t) 
\end{split}
\end{align}
where
\begin{align} 
\begin{split} \label{eq:forward_Euler_rv_1}
u(x_t) &= \nabla \Phi(x_t)
\end{split}
\end{align}
where we denote $x_t$ as the discretizations random variables $x(t)$.

Assuming that the $\alpha_t: \Omega \rightarrow \Omega$ is a diffeomorphism (bijective function with differentiable inverse), the push-forward operator $\alpha_{t\#}: \mathbb{R} \rightarrow \mathbb{R}$ on density function ${\rho}^{\Phi}_t \mapsto {\rho}^{\Phi}_{t+\Delta_t} := \alpha_{t\#} {\rho}^{\Phi}_t$ is defined by:
\begin{align} 
\begin{split} \label{eq:change_of_variables}
\int_{\Omega} {\rho}^{\Phi}_{t+\Delta_t}(x) \, g(x) \; dx
= \int_{\Omega} \alpha_{t\#} {\rho}^{\Phi}_t(x) \, g(x) \; dx
= \int_{\Omega} {\rho}^{\Phi}_t (x) \, g\big(\alpha_t(x)\big) \; dx
\end{split}
\end{align}
for any measurable function $g$.

Associated with the change-of-variables formula (\ref{eq:change_of_variables}) is the following density transformation:
\begin{align} 
\begin{split} \label{eq:density_transformation}
{\rho}^{\Phi}_{t+\Delta_t}\big(\alpha_t(x)\big) &= \frac{1}{|D \alpha_t|} \, {\rho}^{\Phi}_{t}(x)
\end{split}
\end{align}
where $|D \alpha_t|$ denotes the Jacobian determinant of $\alpha_t$. 

From (\ref{eq:homotopy_PDE}) and (\ref{eq:marginalized_homotopy_derivative_expand}), we have
\begin{align} \label{eq:marginalized_homotopy_derivative_repeat}
\begin{split}
\frac{\partial \ln \bar{\rho}_t(x)}{\partial t} &= \frac{1}{\bar{\rho}_t(x)} \, \frac{\partial \bar{\rho}_t(x)}{\partial t} 
= - \, \frac{1}{\bar{\rho}_t(x)} \, \frac{1}{2} \, \mean_{{p}_{\text{data}}(\bar{x})} \Big[\rho_t(x,\bar{x}) \, \big( \gamma(x,\bar{x}) - \bar{\gamma}(x,\bar{x}) \big) \Big]
\end{split}
\end{align}

Applying the forward Euler method to (\ref{eq:marginalized_homotopy_derivative_repeat}), we obtain
\begin{align} 
\begin{split} \label{eq:forward_Euler_homotopy}
\ln \bar{\rho}_{t+\Delta_t}(x) \geq \ln \bar{\rho}_t(x) - \frac{\Delta_t}{2} \, \frac{1}{\bar{\rho}_t(x)} \, \mean_{{p}_{\text{data}}(\bar{x})} \Big[\rho_t(x,\bar{x}) \, \big( \gamma(x,\bar{x}) - \bar{\gamma}(x,\bar{x}) \big) \Big]
\end{split}
\end{align}

Applying the change-of-variables formula (\ref{eq:change_of_variables}) and density transformation (\ref{eq:density_transformation}), then substituting (\ref{eq:forward_Euler_homotopy}) into the KLD $\KLD \big[ {\rho}^{\Phi}_{t + \Delta_t} \,\|\, \bar{\rho}_{t + \Delta_t} \big]$ at time $t + \Delta_t$, we have
\begin{align}
\begin{split} \label{eq:min_KLD_problem_expand}
&\KLD\big[ {\rho}^{\Phi}_{t+\Delta_t}(x) \,\|\, \bar{\rho}_{t+\Delta_t}(x) \big]
= \int_{\Omega} {\rho}^{\Phi}_{t}(x) \, \ln \Bigg( \frac{{\rho}^{\Phi}_{t+\Delta_t}\big(\alpha_t(x)\big)}{\bar{\rho}_{t+\Delta_t}\big(\alpha_t(x)\big)} \Bigg) \; dx \\
&= \int_{\Omega} {\rho}^{\Phi}_{t}(x) \, \bigg( \ln {\rho}^{\Phi}_t(x) - \ln |D \alpha_t| - \ln \bar{\rho}_t\big(\alpha_t(x)\big) \\
&\qquad\qquad\qquad+ \frac{\Delta_t}{2} \, \frac{1}{\bar{\rho}_t\big(\alpha_t(x)\big)} \, \mean_{{p}_{\text{data}}(\bar{x})} \bigg[ \rho_t\big(\alpha_t(x), \bar{x}\big) \, \Big( \gamma\big(\alpha_t(x), \bar{x}\big) - \bar{\gamma}\big(\alpha_t(x), \bar{x}\big) \Big) \bigg] + C \bigg) \; dx \\
\end{split}
\end{align} 

Consider minimizing the KLD (\ref{eq:min_KLD_problem_expand}) with respect to ${\alpha_t}$ as follows:
\begin{align} \label{eq:min_LBKLD_problem}
\begin{split}
&\min_{\alpha_t} \;\; \KLD (\alpha_t) \\
&= \min_{\alpha_t} \;\; \underbrace{\frac{\Delta_t}{2} \int_{\Omega} {\rho}^{\Phi}_{t}(x) \, \frac{1}{\bar{\rho}_t\big(\alpha_t(x)\big)} \, \mean_{{p}_{\text{data}}(\bar{x})} \bigg[ \rho_t\big(\alpha_t(x),\bar{x}\big) \, \Big( \gamma\big(\alpha_t(x), \bar{x}\big) - \bar{\gamma}\big(\alpha_t(x), \bar{x}\big) \Big) \bigg] \; dx}_{\mathcal{D}^\mathrm{KL}_1(\alpha_t)} \\
&\qquad\qquad\!\!\!- \underbrace{\int_{\Omega} {\rho}^{\Phi}_{t}(x) \, \ln \bar{\rho}_t(\alpha_t(x)) \; dx}_{\mathcal{D}^\mathrm{KL}_2(\alpha_t)} 
\,-\, \underbrace{\int_{\Omega} {\rho}^{\Phi}_{t}(x) \, \ln |D \alpha_t| \; dx}_{\mathcal{D}^\mathrm{KL}_3(\alpha_t)} 
\end{split}
\end{align}
where we have neglected the constant terms that do not depend on ${\alpha_t}$.

To solve the optimization (\ref{eq:min_LBKLD_problem}), we consider the following optimality condition in the first variation of $\KLD$:
\begin{align} \label{eq:optimality_P1}
\mathcal{I}(\alpha,\nu) = \frac{d}{d\epsilon} \, \KLD\big(\alpha(x) + \epsilon \, \nu(x)\big) \, \bigg|_{\epsilon=0} = 0
\end{align}
which must hold for all trial function $\nu(x)$. 

Taking the variational derivative of the first functional $\mathcal{D}^\mathrm{KL}_1$ in (\ref{eq:min_LBKLD_problem}), we have
\begin{align} \label{eq:I_1_full}
\begin{split}
&\mathcal{I}^{1}(\alpha,\nu) = \frac{d}{d\epsilon} \, \mathcal{D}^\mathrm{KL}_1(\alpha + \epsilon \nu) \, \bigg|_{\epsilon=0} \\
&= \frac{\Delta}{2} \int_{\Omega} {\rho}^{\Phi}(x) \, \frac{d}{d\epsilon} \bigg\{ \frac{1}{\bar{\rho}(\alpha + \epsilon \nu)} \, \mean_{{p}_{\text{data}}(\bar{x})} \Big[ \rho(\alpha + \epsilon \nu, \bar{x}) \, \big( \gamma(\alpha + \epsilon \nu, \bar{x}) - \bar{\gamma}(\alpha + \epsilon \nu, \bar{x}) \big) \Big] \bigg\} \, \Bigg|_{\epsilon=0} \; dx \\
&= \frac{\Delta}{2} \int_{\Omega} {\rho}^{\Phi}(x) \, D \, \bigg\{ \frac{1}{\bar{\rho}(x)} \, \mean_{{p}_{\text{data}}(\bar{x})} \Big[ \rho(x, \bar{x}) \, \big( \gamma(x, \bar{x}) - \bar{\gamma}(x, \bar{x}) \big) \Big] \bigg\} \; \nu \; dx
\end{split}
\end{align}
where $D g := \nabla^T g$ denotes the Jacobian of function $g(x)$ with respect to $x$.

A Taylor series expansion of the derivative $\frac{\partial g}{\partial x_i}(\alpha)$ with respect to $x_i$ yields
\begin{align}
\begin{split} \label{eq:Taylor_series_expansion_2}
\frac{\partial g(\alpha)}{\partial x_i} &= \frac{\partial g(x + \Delta u)}{\partial x_i}
= \frac{\partial g(x)}{\partial x_i} + \Delta \, \sum_{j} \frac{\partial^2 g(x)}{\partial x_i \, \partial x_j} \, u_{j} + O(\Delta^2) \\
\end{split}
\end{align}

Using the Taylor series expansion (\ref{eq:Taylor_series_expansion_2}), (\ref{eq:I_1_full}) can be written in tensor notation as follows:
\begin{align} \label{eq:I_1_full_expand}
\begin{split}
\mathcal{I}^{1}(\alpha,\nu) 
&= \frac{\Delta}{2} \int_{\Omega} {\rho}^{\Phi}(x) \, \sum_{i} \, \frac{\partial}{\partial x_i} \bigg\{ \frac{1}{\bar{\rho}(x)} \, \mean_{{p}_{\text{data}}(\bar{x})} \Big[ \rho(x; \bar{x}) \, \big( \gamma(x,\bar{x}) - \bar{\gamma}(x,\bar{x}) \big) \Big] \bigg\} \, \nu_{i} \; dx \;+\; O(\Delta^2)
\end{split}
\end{align}



Taking the variational derivative of the second functional $\mathcal{D}^\mathrm{KL}_2$ in (\ref{eq:min_LBKLD_problem}) yields
\begin{align}
\begin{split} \label{eq:I_2_full}
\mathcal{I}^{2}(\alpha,\nu) &= \frac{d}{d\epsilon} \, \mathcal{D}^\mathrm{KL}_2(\alpha + \epsilon \nu) \, \bigg|_{\epsilon=0} \\
&= \int_{\Omega} {\rho}^{\Phi}(x) \, \frac{d}{d\epsilon} \ln \bar{\rho}(\alpha + \epsilon \nu) \, \bigg|_{\epsilon=0} \; dx \\
&= \int_{\Omega} {\rho}^{\Phi}(x) \, \frac{1}{\bar{\rho}(\alpha)} \, \nabla \bar{\rho}(\alpha) \cdot \nu \; dx \\
&= \int_{\Omega} {\rho}^{\Phi}(x) \, \nabla \ln \bar{\rho}(\alpha) \cdot \nu \; dx \\
\end{split}
\end{align}
where we have used the derivative identity $d \ln g = \frac{1}{g} \, d g$ to obtain the second equation.

Using the Taylor series expansion (\ref{eq:Taylor_series_expansion_2}), (\ref{eq:I_2_full}) can be written in tensor notation as follows:
\begin{align} \label{eq:I_2_full_expand}
\begin{split}
\mathcal{I}^{2}(\alpha,\nu) &= - \int_{\Omega} {\rho}^{\Phi}(x) \, \sum_{i} \Bigg( \frac{\partial \ln \bar{\rho}(x)}{\partial x_i}
- \Delta \, \sum_{j} \, \frac{\partial^2 \ln \bar{\rho}(x)}{\partial x_i \, \partial x_j} \, u_{j} \Bigg) \, \nu_{i} \; dx \;+\; O(\Delta^2) \\
&= - \int_{\Omega} {\rho}^{\Phi}(x) \, \sum_{i} \Bigg( \frac{\partial \ln \bar{\rho}(x)}{\partial x_i}
- \Delta \, \sum_{j} \, \frac{\partial^2 \ln \bar{\rho}(x)}{\partial x_i \, \partial x_j} \, u_{j} \Bigg) \, \nu_{i} \; dx \;+\; O(\Delta^2)
\end{split}
\end{align}

Similarly, taking the variational derivative of the $\mathcal{D}^\mathrm{KL}_3$ term in (\ref{eq:min_LBKLD_problem}), we have
\begin{align}
\begin{split} \label{eq:I_3_full}
\mathcal{I}^{3}(\alpha,\nu) &= \frac{d}{d\epsilon} \, \mathcal{D}^\mathrm{KL}_3(\alpha + \epsilon \nu) \, \bigg|_{\epsilon=0} \\
&= \int_{\Omega} {\rho}^{\Phi}(x) \, \frac{d}{d\epsilon} \ln \big| D (\alpha + \epsilon \nu) \big| \, \bigg|_{\epsilon=0} \; dx \\
&= \int_{\Omega} {\rho}^{\Phi}(x) \, \frac{1}{\big|D\alpha\big|} \, \frac{d}{d\epsilon} \big| D(\alpha + \epsilon \nu) \big| \, \bigg|_{\epsilon=0} \; dx \\
&= \int_{\Omega} {\rho}^{\Phi}(x) \, \tr \big({D\alpha}^{-1} D\nu\big) \; dx
\end{split}
\end{align}
where we have used the following Jacobi's formula:
\begin{align} \label{eq:Jacobi_formula}
\frac{d}{d\epsilon} \big|D (\alpha + \epsilon \nu) \big| \, \bigg|_{\epsilon=0} = \left|D\alpha\right| \tr\left(D\alpha^{-1}D\nu\right)
\end{align}
to obtain the last equation in (\ref{eq:I_3_full}).

The inverse of Jacobian $D\alpha^{-1}$ can be expanded via Neuman series to obtain
\begin{align}
\begin{split} \label{eq:Neuman_series}
D\alpha^{-1} &= \big(\, \mathrm{I} + \Delta \, Du \,\big)^{-1} 
= \mathrm{I} - \Delta \, Du \;+\; O(\Delta^2)
\end{split}
\end{align}

Substituting in (\ref{eq:Neuman_series}) and using the Taylor series expansion (\ref{eq:Taylor_series_expansion_2}), (\ref{eq:I_2_full}) can be written in tensor notation as follows:
\begin{align} \label{eq:I_3_full_expand}
\begin{split} 
\mathcal{I}^{3}(\alpha,\nu) &= \int_{\Omega} \sum_{i} \Bigg( {\rho}^{\Phi}(x) \, \frac{\partial \nu_{i}}{\partial x_{i}} - \Delta \sum_{j} \, {\rho}^{\Phi}(x) \, \frac{\partial u_{j}}{\partial x_{i}} \, \frac{\partial \nu_{i}}{\partial x_{j}} \Bigg) \; dx \,+\, O(\Delta^2) \\
&= \int_{\Omega} \sum_{i} \Bigg( \frac{\partial {\rho}^{\Phi}(x)}{\partial x_{i}} \, \nu_{i} - \Delta \sum_{j} \, \frac{\partial}{\partial x_{j}} \bigg\{ {\rho}^{\Phi}(x) \, \frac{\partial u_{j}}{\partial x_{i}} \bigg\} \, \nu_{i} \Bigg) \; dx \,+\, O(\Delta^2) \\
&= \int_{\Omega} \sum_{i} \Bigg( \frac{\partial {\rho}^{\Phi}(x)}{\partial x_{i}} - \Delta \sum_{j} \, \frac{\partial}{\partial x_{j}} \bigg\{ {\rho}^{\Phi}(x) \, \frac{\partial u_{j}}{\partial x_{i}} \bigg\} \Bigg) \, \nu_{i} \; dx \,+\, O(\Delta^2)
\end{split}
\end{align}
where we have used integration by parts to obtain the second equation.

Taking the limit $\lim \Delta \rightarrow 0$, the terms $O(\Delta^2)$ that approach zero exponentially vanish.
Subtracting (\ref{eq:I_1_full_expand}) by (\ref{eq:I_2_full_expand}) and (\ref{eq:I_3_full_expand}) then equating to zero, we obtain the first-order optimality condition (\ref{eq:optimality_P1}) as follows:
\begin{align} \label{eq:sum_of_derivatives}
\begin{split}
\int_{\Omega} \bar{\rho}(x) \, \sum_{i} \Bigg( \sum_{j} &-\, \frac{\partial}{\partial x_{i}} \bigg\{ \frac{1}{\bar{\rho}(x)} \, \frac{\partial}{\partial x_{j}} \Big\{ \bar{\rho}(x) \, u_{j} \Big\} \bigg\} \\
&+ \frac{1}{2} \, \frac{\partial}{\partial x_i} \bigg\{ \frac{1}{\bar{\rho}(x)} \, \mean_{{p}_{\text{data}}(\bar{x})} \Big[ \rho(x;\bar{x}) \, \big( \gamma(x, \bar{x}) - \bar{\gamma}(x, \bar{x}) \big) \Big] \bigg\} \Bigg) \; \nu_{i} \; dx = 0
\end{split}
\end{align}
where we have assumed that ${\rho}^{\Phi}(x) \equiv \bar{\rho}(x)$ holds, and used the following identities:
\begin{align} \label{eq:used_identities_1}
\begin{split}
\frac{\partial \ln \bar{\rho}(x)}{\partial x_i} &= \frac{1}{\bar{\rho}(x)} \, \frac{\partial \bar{\rho}(x)}{\partial x_{i}} \\
\frac{\partial^2 \ln \bar{\rho}(x)}{\partial x_i \, \partial x_j} &= \frac{\partial}{\partial x_i} \bigg(\frac{1}{\bar{\rho}(x)} \frac{\partial \bar{\rho}(x)}{\partial x_j}\bigg)
\end{split}
\end{align}

Given that $\nu_{i}$ can take any value, the equation (\ref{eq:sum_of_derivatives}) holds (in the weak sense) only if the terms within the round bracket vanish. 
Integrating this term with respect to the $x_i$, we are left with
\begin{align}
\begin{split} \label{eq:weak_sense_integrated}
\sum_{j} \, \frac{\partial}{\partial x_{j}} \Big\{ \bar{\rho}(x) \, u_{j} \Big\} 
= \frac{1}{2} \, \mean_{{p}_{\text{data}}(\bar{x})} \Big[ \rho(x;\bar{x}) \, \big( \gamma(x, \bar{x}) - \bar{\gamma}(x, \bar{x}) \big) \Big] \,+\, \bar{\rho}(x) \, C 
\end{split}
\end{align}
which can also be written in vector notation as follows:
\begin{align} \label{eq:weak_solution_PDE_full}
\begin{split} 
& \nabla \cdot \big( \bar{\rho}(x) \, u \big) = \frac{1}{2} \, \mean_{{p}_{\text{data}}(\bar{x})} \Big[ \rho(x;\bar{x}) \, \big( \gamma(x, \bar{x}) - \bar{\gamma}(x, \bar{x}) \big) \Big] \,+\, \bar{\rho}(x) \, C
\end{split}
\end{align}

To find the scalar constant $C$, we integrate both sides of (\ref{eq:weak_solution_PDE_full}) to get
\begin{align} \label{eq:integration_by_parts_full}
\begin{split} 
\int_{\Omega} \nabla \cdot \big( \bar{\rho}(x) \, u \big) \; dx 
&= \frac{1}{2} \, \int_{\Omega} \mean_{{p}_{\text{data}}(\bar{x})} \Big[ \rho(x;\bar{x}) \, \big( \gamma(x, \bar{x}) - \bar{\gamma}(x,\bar{x}) \big) \Big] \; dx \,+\, \int_{\Omega} \bar{\rho}(x) \, C \; dx \\
&= \frac{1}{2} \, \int_{\Omega} \mean_{{p}_{\text{data}}(\bar{x})} \Big[ \rho(x;\bar{x}) \, \big( \gamma(x, \bar{x}) - \bar{\gamma}(x,\bar{x}) \big) \Big] \; dx \,+\, C
\end{split}
\end{align}

Applying the divergence theorem to the left-hand side of (\ref{eq:integration_by_parts_full}), we have 
\begin{align} \label{eq:divergence_theorem}
\begin{split} 
\int_{\Omega} \nabla \cdot \big( \bar{\rho}(x) \, u \big) \; dx 
= \int_{\partial \Omega} \bar{\rho}(x) \, u \cdot \hat{n} \; dx
\end{split}
\end{align}
where $\hat{n}$ is the outward unit normal vector to the boundary $\partial \Omega$ of $\Omega$.

Given that $\bar{\rho}(x)$ is a normalized (proper) density with compact support (vanishes on the boundary), the term (\ref{eq:divergence_theorem}) becomes zero and we obtain $C = 0$.
Substituting this and $u(x) = \nabla \Phi_{\theta}(x)$ into (\ref{eq:weak_solution_PDE_full}), we arrive at the PDE
\begin{align} \label{eq:Proposition_1_PDE_general}
\begin{split} 
\nabla \cdot \big( \bar{\rho}_t(x) \, \nabla \Phi(x) \big) = \frac{1}{2} \, \mean_{{p}_{\text{data}}(\bar{x})} \Big[ \rho_t(x,\bar{x}) \, \big( \gamma(x,\bar{x}) - \bar{\gamma}(x,\bar{x}) \big) \Big]
\end{split}
\end{align}

Assume that the base case ${\rho}^{\Phi}_0(x) \equiv \bar{\rho}_0(x)$ holds, and that there exists a solution to (\ref{eq:Proposition_1_PDE_general}) for every $t$. The proposition follows by the principle of induction.


\end{proof}


\subsection{Proof of Proposition \ref{thm:proposition_3}} \label{Appendix:C}

\begin{proof} \label{pf:proposition_2}
The energy loss function in (\ref{eq:variational_functional}) can generally be written as follows:
\begin{align} \label{eq:variational_functional_rewritten}
\begin{split}
\mathcal{L}(\Phi, t) = &\ \frac{1}{2} \, \mean_{\rho(x;\bar{x},t) \, {p}_{\text{data}}(x)} \Big[ \Phi(x) \, \big( \gamma(x,\bar{x}) - \bar{\gamma}(x,\bar{x}) \big) \Big] 
+ \frac{1}{2} \, \mean_{\bar{\rho}(x;t)} \, \Big[ \big\| \nabla \Phi(x) \big\|^{2} \Big]
\end{split}
\end{align}
where we have assumed, without loss of generality, that a normalized potential energy $\bar{E}_{\theta}(x;t) = 0$. For an unnormalized solution $\Phi(x)$, we can always obtain the desired normalization by subtracting its mean.

The optimal solution $\Phi$ of the functional (\ref{eq:variational_functional_rewritten}) is given by the first-order optimality condition:
\begin{align} \label{eq:optimality_P2}
\mathcal{I}(\Phi,\Psi) &= \frac{d}{d\epsilon} \, \mathcal{L}(\Phi(x) + \epsilon \Psi(x), t) \, \bigg|_{\epsilon=0} = 0
\end{align}
which must hold for all trial function $\Psi$.

Taking the variational derivative of the particle flow objective (\ref{eq:optimality_P2}) with respect to $\epsilon$, we have
\begin{align} \label{eq:optimality_P2_full}
\begin{split}
&\mathcal{I}(\Phi,\Psi) = \frac{d}{d\epsilon} \, \mathcal{L}(\Phi + \epsilon \Psi) \, \bigg|_{\epsilon=0} \\
&= \frac{1}{2} \, \int_{\Omega \times \Omega} {p}_{\text{data}}(x) \, \rho(x;\bar{x}) \, \big( \gamma(x,\bar{x}) - \bar{\gamma}(x,\bar{x}) \big) \, \frac{d}{d\epsilon} (\Phi + \epsilon \Psi) \; d\bar{x} \, dx \\
&\quad\,+ \frac{1}{2} \, \int_{\Omega} \bar{\rho}(x) \, \frac{d}{d\epsilon} \big\| \nabla (\Phi + \epsilon \Psi) \big\|^2 \; dx
\\
&= \frac{1}{2} \, \int_{\Omega \times \Omega} {p}_{\text{data}}(x) \, \rho(x;\bar{x}) \, \big( \gamma(x,\bar{x}) - \bar{\gamma}(x,\bar{x}) \big) \, \Psi \; d\bar{x} \, dx
\;+\; \int_{\Omega} \bar{\rho}(x) \, \nabla \Phi \cdot \nabla \Psi \; dx
\end{split}
\end{align}

Given that $\Phi \in \mathcal{H}^1_0(\Omega; \bar{\rho})$, the energy values vanish on the boundary $\partial \Omega$. Therefore, the second summand of the last expression in (\ref{eq:optimality_P2_full}) can be written, via multivariate integration by parts, as
\begin{align} \label{eq:integration by parts}
\begin{split}
&\int_{\Omega} \bar{\rho}(x) \, \nabla \Phi \cdot \nabla \Psi = - \int_{\Omega} \nabla \cdot \big( \bar{\rho}(x) \, \nabla \Phi \big) \, \Psi \; dx
\end{split}
\end{align}

By substituting (\ref{eq:integration by parts}) into (\ref{eq:optimality_P2_full}), we get
\begin{align} \label{eq:optimality_P2_full_simplify}
\begin{split}
\mathcal{I}(\Phi,\Psi) &= \frac{1}{2} \, \int_{\Omega} \, \int_{\Omega} {p}_{\text{data}}(x) \rho(x;\bar{x}) \, \big( \gamma(x,\bar{x}) - \bar{\gamma}(x,\bar{x}) \big) \, \Psi \; d\bar{x} \, dx
\;-\; \int_{\Omega} \nabla \cdot \big( \bar{\rho}(x) \, \nabla \Phi \big) \, \Psi \; dx \\ 
&= \int_{\Omega} \bigg( \, \frac{1}{2} \, \int_{\Omega} {p}_{\text{data}}(x) \rho(x;\bar{x}) \, \big( \gamma(x,\bar{x}) - \bar{\gamma}(x,\bar{x}) \big) \; d\bar{x}
\,-\, \int_{\Omega} \nabla \cdot \big( \bar{\rho}(x) \, \nabla \Phi \big) \bigg) \Psi  \; dx
\end{split}
\end{align}

and equating it to zero, we obtain the weak formulation (\ref{eq:weak_formulation}) of the probabilistic Poisson's equation.

Given that the Poincaré inequality (\ref{eq:Poincare_inequality}) holds, \cite[Theorem~2.2]{Laugesen} presents a rigorous proof of existence and uniqueness for the solution $\Phi$ to the weak formulation (\ref{eq:weak_formulation}), based on the Hilbert-space form of the Riesz representation theorem.
\end{proof}

\subsection{Derivation of Time-varying Mean and Variance in (\ref{eq:unmarginalized_homotopy_statistics})} \label{Appendix:D}

Given the following marginal Gaussian distribution for $z$ and a conditional Gaussian distribution for $x$ given $x$, as defined in Section \ref{ssect:LogHomotopy}:
\begin{subequations} \label{eq:Gaussian_distributions}
\begin{align}
\label{eq:marginal_Gaussian}
&q(x) = \mathcal{N}(x; 0, \Lambda) \\
\label{eq:conditional_Gaussian}
&p(\bar{x}|x) = \mathcal{N}(\bar{x}; x, \Pi)
\end{align}
\end{subequations}

The posterior distribution of $x$ given $\bar{x}$ is obtained via Bayes' theorem as
\begin{align} \label{eq:Bayes_theorem}
\begin{split}
p(x|\bar{x}) = \frac{p(\bar{x}|x) \, q(x)}{\int_{\Omega} p(\bar{x}|x) \, q(x) \; dx} = \mathcal{N}(x; \mu, \Sigma)
\end{split}
\end{align}
and remains a Gaussian, whose mean and variance are given by:
\begin{subequations} \label{eq:unmarginalized_homotopy_statistics_derivation}
\begin{align}
&\mu(\bar{x}) = \Sigma \, \Pi^{-1} \, \bar{x} \\
&\Sigma = \big(\Lambda^{-1} \,+\, \Pi^{-1}\big)^{-1}
\end{align}
\end{subequations}

In fact, the conditional homotopy (\ref{eq:unmarginalized_homotopy_h}) can be written as
\begin{align} \label{eq:unmarginalized_homotopy_Bayes}
\begin{split}
\rho(x; \bar{x}, t) 
= \frac{p(\bar{x};t|x) \, q(x)}{\int_{\Omega} p(\bar{x};t|x) \, q(x) \; dx}
\end{split}
\end{align}
where 
\begin{align} \label{eq:time_varying_conditional_likelihood}
\begin{split}
p(\bar{x};t|x) = \mathcal{N}(x; \mu, \frac{1}{t} \Pi)
\end{split}
\end{align}
Notice that the terms involving $t$ in the numerator and denominator of (\ref{eq:unmarginalized_homotopy_Bayes}) cancel each other out.
Substituting the variance of (\ref{eq:time_varying_conditional_likelihood}) into (\ref{eq:unmarginalized_homotopy_Bayes}) and using (\ref{eq:Bayes_theorem})-(\ref{eq:unmarginalized_homotopy_statistics_derivation}), we obtain (\ref{eq:unmarginalized_homotopy_statistics}).

\section{Experimental Details}
\label{sec:experimental_details}

\subsection{Model architecture}
Our network structure is based on the Wide ResNet \cite{wideresnet}. We adopt the same model hyperparameters used in \cite{Diffusion_Recovery} for different datasets. In particular, the number of downsampled resolutions increases with the image size of the dataset and the number of ResBlocks in each resolution varies. Nevertheless, there are a few major differences between our network model and the ones used in \cite{Diffusion_Recovery}: 
\begin{enumerate}
\item We replace LeakyReLU activations with Gaussian Error Linear Unit (GELU) activations \cite{GELU}, which we found 
improves training stability and convergence.
\item We do not use spectral normalization \cite{Miyato}; instead, we use weight normalization with data-dependent initialization \cite{Salimans}. 
\item Following \cite{VAEBM, NVAE}, our training includes an additional spectral regularization loss which penalizes the spectral norm of each convolutional layer in the Wide ResNet to regularize the sharpness of the energy model.
\item We drop the time embedding to render the time variable implicit in our energy model.
\end{enumerate}



\subsection{Training}
We use the Lamb optimizer \cite{LAMB} and a learning rate of $0.001$ for all the experiments. We find that Lamb performs better than Adam over large learning rates. Following \cite{Diffusion_Recovery}, we set a smaller $\beta_1$ of $0.5$ in Lamb for the more high-resolution CelebA and LSUN images to help stabilize training.
For CIFAR-10, CelebA and LSUN, we use a batch size of 256, 128 and 64, respectively.
For all experiments, we set a spectral gap constant $\lambda$ of $0.001$, and a sharpness constant $\varepsilon$ of $0.0001$ in our training. Here, we set the standard deviation $\omega$ of the prior density to be $1$ so that the data likelihood homotopy is variance-preserving. Also, we set the standard deviation $\sigma$ of conditional data likelihood to be $0.01$ so that the difference between samples $x$ and data $\bar{x}$ is indistinguishable to human eyes \cite{NCSN}.
All models are trained for 1.3M iterations on a single NVIDIA A100 (80GB) GPU.

\subsection{Numerical Solver}
In our experiments, the default solver of ODEs used is the black box solver in the Scipy library with the RK45 method \cite{RK45} following \cite{PFGM}. Since time variable $t$ is implicit in our energy model $\Phi(x(t))$, we can set a longer ODE interval, allowing the additional ODE iterations to further refine the samples within regions of high likelihood and improve the quality of generated images. We observe that setting a terminal time $t_\text{end}$ of $1.625$ for the RK45 ODE solver gives the best results.

\subsection{Datasets}
We use the following datasets in our experiments: CIFAR-10 \cite{cifar10}, CelebA \cite{celeba} and LSUN \cite{lsun}. CIFAR-10 is of resolution $32 \times 32$, and contains $50,000$ training images and $10,000$ test images. CelebA contains $202,599$ face images, of which $162,770$ are training images and $19,962$ are test images. For processing, we first clip each image to $178 \times 178$ and then resize it to $64 \times 64$. For LSUN, we use the church outdoor and bedroom categories, which contain $126,227$ and $3,033,042$ training images respectively. Both categories contain $300$ test images. For processing, we first crop each image to a square image whose side is of length which is the minimum of 
the height and weight, and then we resize it to $64 \times 64$ or $128 \times 128$. For resizing, we set the anti-alias to True. We apply horizontal random flip as data augmentation for all datasets during training.

\subsection{Quantitative Evaluation}
We employ the FID and inception scores as quantitative evaluation metrics for assessing the quality of generated samples. For CIFAR-10, we compute the Frechet distance between $50,000$ samples and the pre-computed statistics on the training set [13]. For CelebA $64 \times 64$, we follow the setting in \cite{NCSNv2} where the distance is computed between $5,000$ samples and the 
pre-computed statistics on the test set. For model selection, we follow \cite{NCSN++} and pick the checkpoint with the smallest FID scores, computed on 2,500 samples every 10,000 iterations.

\subsection{Potential Societal Impact}
Generative models is a rapidly growing field of study with overarching implications in science and society. Our work proposes a new generative model \textsf{VAPO} that allows image generation via efficient and adaptive sampling. 
The usage of the proposed model could have both positive and negative outcomes depending on the downstream application. For example, \textsf{VAPO} can be used to efficiently produce high-quality image/audio samples via the fast backward ODE. At the same time, it could promote deepfake technology and undermine social security. Generative models are also vulnerable to backdoor adversarial attacks on publicly available training data. Addressing the above concerns requires further collaborative research efforts aimed at mitigating misuse of AI technology.

\section{Mode Evaluation} \label{sec:mode_evaluation}
In this section, we evaluate the mode coverage and over-fitting of the proposed \textsf{VAPO}.

\subsection{Model Over-fitting and Generalization}
\begin{figure}[ht]
\centering
\includegraphics[width=1.0\columnwidth]{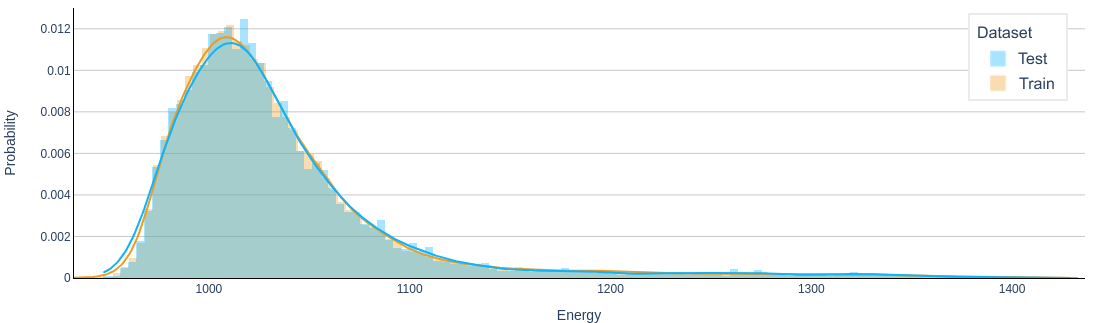}
\caption{Histogram of energy output for CIFAR-10 train and test set.}
\label{fig:histogram_cifar10}
\end{figure}

To 
assess over-fitting, Figure \ref{fig:histogram_cifar10} plots the histogram of the energy outputs on the CIFAR-10 train and test dataset. The energy histogram shows that the learned energy model assigns similar energy values to both train and test set images. This indicates that \textsf{VAPO} generalizes well to unseen test data and extensively covers all the modes in the training data. 

\begin{figure}[ht]
\centering
\includegraphics[width=1.0\columnwidth]{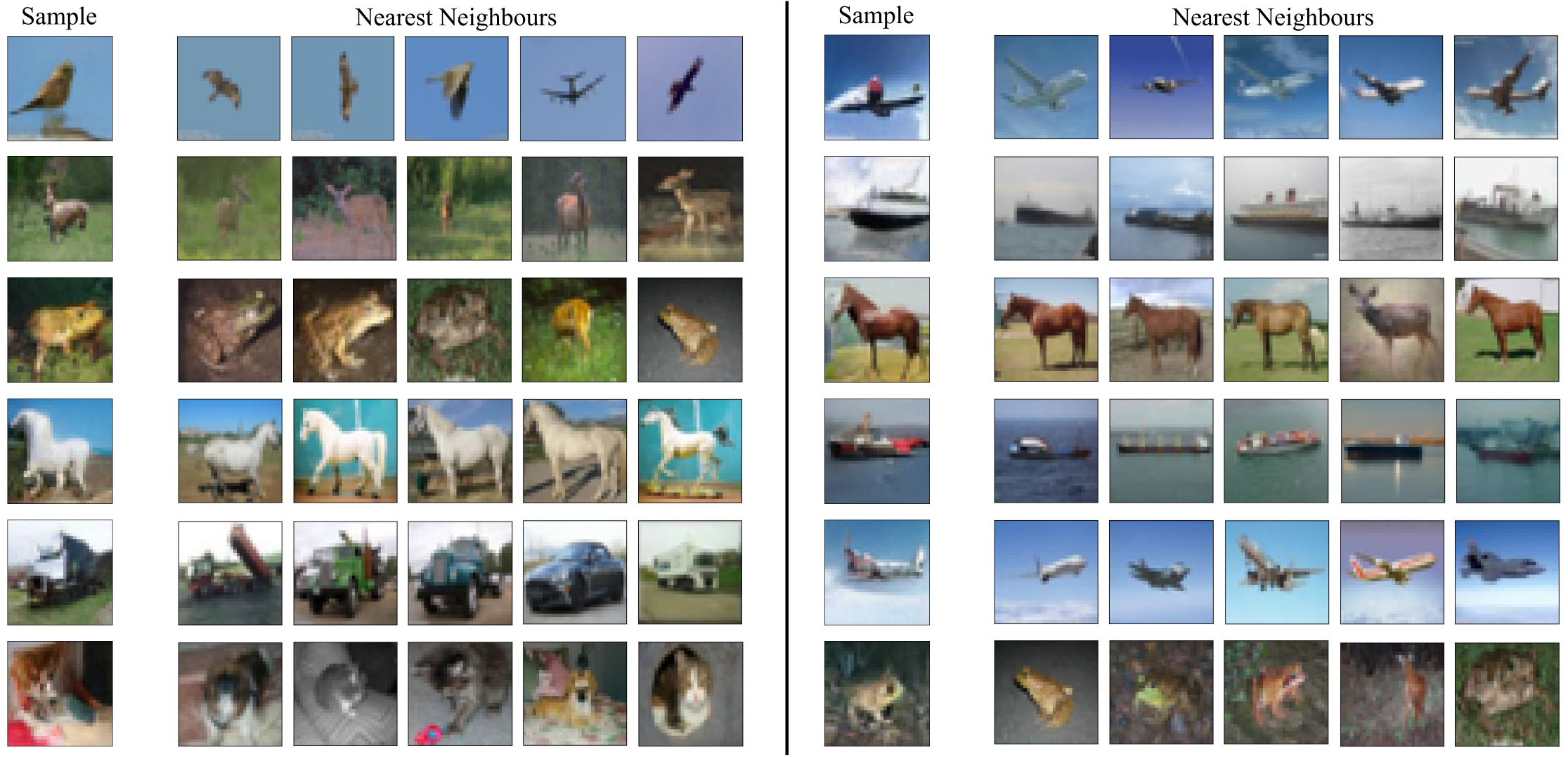}
\caption{Generated samples and their five nearest neighbours in the CIFAR-10 train set based on pixel distance.}
\label{fig:knn_cifar10}
\end{figure}

In addition, Figure \ref{fig:knn_cifar10} presents the nearest neighbors of the generated samples in the train set of CIFAR-10. It shows that nearest neighbors are significantly different from the generated samples, thus suggesting that our models do not over-fit the training data and generalize well across the underlying data distribution.

\subsection{Out-of-Distribution Detection}

We evaluate robustness of our proposed \textsf{VAPO} model to anomalous data by assessing its performance on unsupervised out-of-distribution (OOD) detection. Given that potential energy is conjugate to the approximate data likelihood, the energy model can be used to distinguish between the in-distribution and out-distribution samples based on the energy values it assigns.
In particular, the energy model trained on CIFAR-10 train set is used for assigning normalized energy values to in-distribution samples (CIFAR-10 test set) and out-distribution samples from various other image datasets. The area under the receiver operating characteristic curve (AUROC) is used as a quantitative metric to determine the efficacy of the \textsf{VAPO} model in OOD detection, where a high AUROC score indicates that the model correctly assigns low energy to out-distribution samples. 

Table \ref{tab:auroc} compares the AUROC scores of \textsf{VAPO} with various likelihood-based and EBM-based models. The result shows that \textsf{VAPO} performs exceptionally well on the CIFAR-10 interpolated dataset. However, its performance is average on CIFAR-100 and SVHN. This suggests that the perturbation of training data using the data likelihood homotopy may not sufficiently explore the data space in comparison to MCMC methods. The investigation into the underlying cause is left for future work.

\begin{table}[ht]
\centering
\caption{Comparison of AUROC scores $\uparrow$ for OOD detection on several datasets.}
\label{tab:auroc}
\begin{tabular}{lrrr}
\toprule
\textbf{Models} & \begin{tabular}[c]{l} \textbf{CIFAR-10}\\\textbf{interpolation}\end{tabular} & \textbf{CIFAR-100} & \textbf{SVHN} \\ \midrule
PixelCNN & 0.71 & 0.63 & 0.32 \\
GLOW & 0.51 & 0.55 & 0.24 \\
NVAE & 0.64 & 0.56 & 0.42 \\
EBM-IG & 0.70 & 0.50 & 0.63 \\
VAEBM & 0.70 & 0.62 & 0.83 \\
CLEL & 0.72 & 0.72 & 0.98 \\
DRL & - & 0.44 & 0.88 \\ \midrule
\textsf{VAPO} (Ours) & 0.78 & 0.50 & 0.61 \\ \bottomrule
\end{tabular}
\end{table}

\section{Additional Results} \label{sec:additional_results}
\begin{figure}[p]
\centering
\includegraphics[width=\columnwidth]{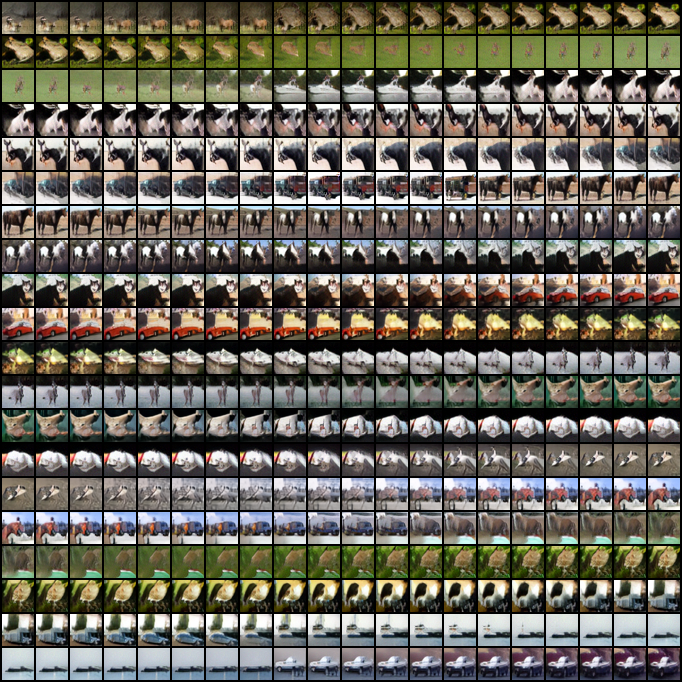}
\caption{Additional interpolation results on unconditional CelebA $64 \times 64$.}
\label{fig:cifar10_interp_big}
\end{figure}

\begin{figure}[p]
\centering
\includegraphics[width=\columnwidth]{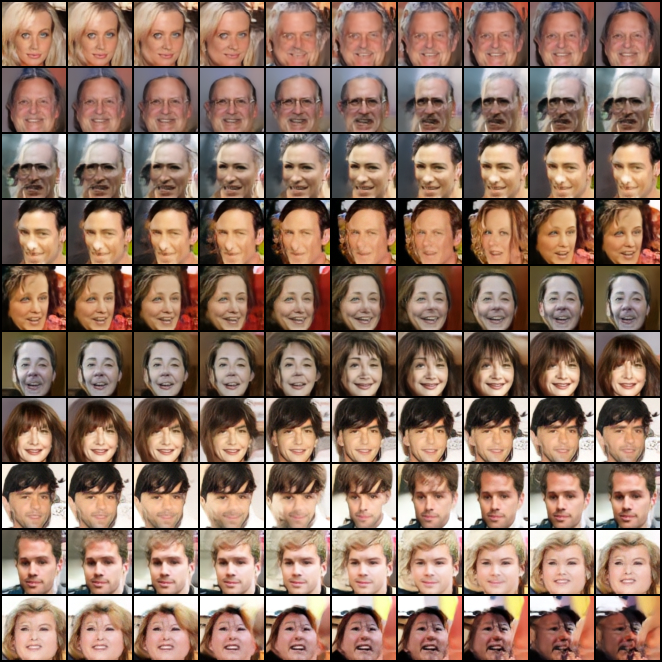}
\caption{Additional interpolation results on unconditional CelebA $64 \times 64$.}
\label{fig:celeba_interp_big}
\end{figure}

\begin{figure}[p]
\centering
\includegraphics[width=\columnwidth]{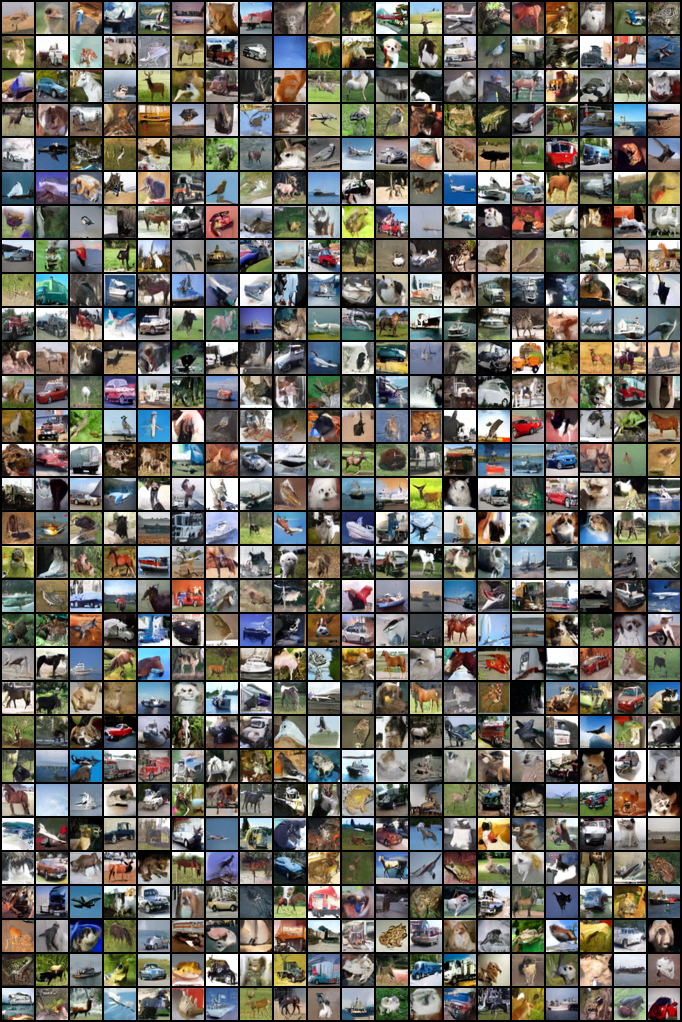}
\caption{Additional uncurated samples on unconditional CIFAR-10 $32 \times 32$.}
\label{fig:cifar10_big}
\end{figure}

\begin{figure}[p]
\centering
\includegraphics[width=\columnwidth]{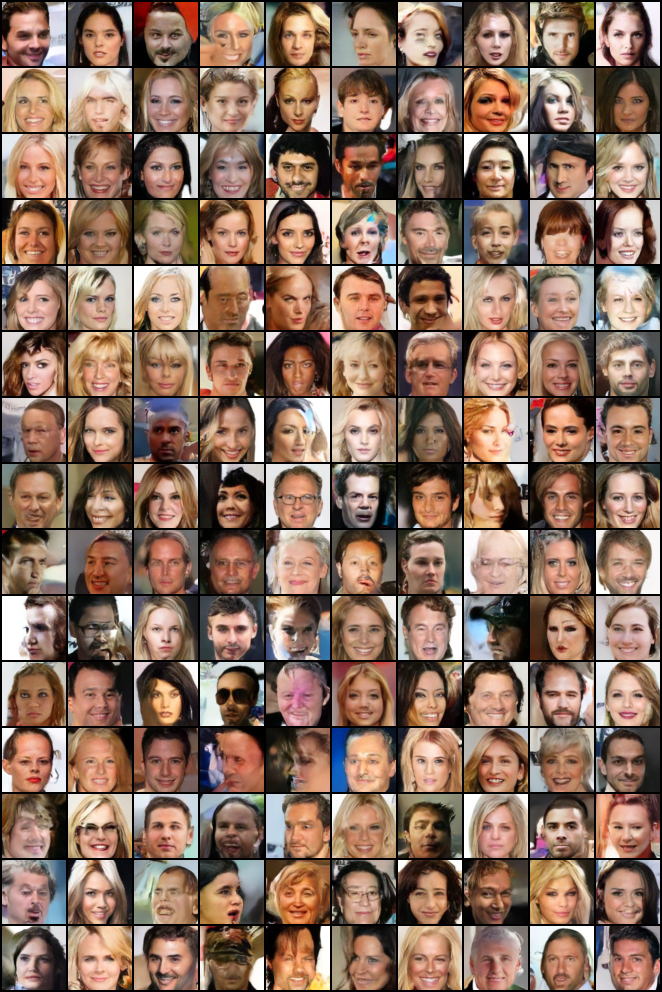}
\caption{Additional uncurated samples on unconditional CelebA $64 \times 64$.}
\label{fig:celeba_big}
\end{figure}

Figures \ref{fig:cifar10_interp_big} and \ref{fig:celeba_interp_big} show additional examples of image interpolation on CIFAR-10 and CelebA $64 \times 64$, respectively.
Figures \ref{fig:cifar10_big} and \ref{fig:celeba_big} show additional uncurated examples of unconditional image generation on CIFAR-10 and CelebA $64 \times 64$, respectively.

\end{document}